\documentclass[final,1p,times,authoryear]{elsarticle}

\newcommand{\ms}[2]{$#1{\scriptstyle\pm#2}$}   

\usepackage{amsmath,amsfonts,amssymb}
\usepackage{booktabs}
\usepackage{multirow}
\usepackage{graphicx}
\usepackage{xcolor}
\usepackage{siunitx}
\usepackage{hyperref}
\usepackage{url}
\usepackage{tabularx}
\usepackage{makecell}
\usepackage{placeins}

\journal{ISPRS Journal of Photogrammetry and Remote Sensing}

\begin{document}

\begin{frontmatter}

\title{Data-Efficient ConvNeXt Pretraining for High-Resolution Sentinel-1 GRD SAR Representation Learning\tnoteref{t1}}

\tnotetext[t1]{This work extends our preliminary conference paper
to be presented at SPIE Asia Pacific Remote Sensing 2026~\citep{natsume2026sparksarspie}
by adding a data-scaling study, a CKA analysis,
a full geographic de-collision experiment, a label efficiency
analysis, a Flood-S1 ceiling ablation, expanded negative results,
and extended methodological detail.}

\author[ccds]{Yuki~Natsume\corref{cor1}}
\ead{nats0001@e.ntu.edu.sg}
\author[eosrs,ase,eee]{Sang-Ho~Yun}
\author[ccds]{Chai~Kiat~Yeo}

\cortext[cor1]{Corresponding author.}

\address[ccds]{College of Computing and Data Science, Nanyang Technological University, Singapore}
\address[eosrs]{Earth Observatory of Singapore, Nanyang Technological University, Singapore}
\address[ase]{Asian School of the Environment, Nanyang Technological University, Singapore}
\address[eee]{School of Electrical and Electronic Engineering, Nanyang Technological University, Singapore}

\begin{abstract}
Labeled Sentinel-1 Ground Range Detected (S1GRD) datasets remain
scarce, limiting supervised deep learning for land-surface remote
sensing, while self-supervised pretraining for this domain has been
dominated by Vision Transformer (ViT)-based masked image modeling
requiring substantial data and compute. We propose SparK-SAR, a
sparse convolutional masked pretraining framework built on SparK and
ConvNeXt, designed to exploit convolutional inductive biases for
S1GRD imagery under constrained compute. SparK-SAR makes three core
technical contributions:
(i) ConvNeXt-RS, a domain-adapted encoder that halves
the stem stride ($4{\times}{\to}2{\times}$) to produce $14{\times}14$
feature maps at stride $16{\times}$, providing $4{\times}$ more
spatial tokens at identical parameter count;
(ii) a lightweight ConvNeXt-512 reconstruction decoder with PixelShuffle 
upsampling that reduces pretraining VRAM by 7.5--8.5\% per GPU relative 
to a UNet-128 baseline while matching its accuracy;
and (iii) an empirical characterization of FP16 mixed-precision pretraining 
instability in the standard SparK UNet-768 decoder on S1GRD data, complemented 
by a finding that finetuning with layer-wise learning rate decay (LLRD) 
improves downstream performance of pretrained encoders. 
On four Copernicus-Bench S1GRD tasks, 
our finetuned 27.8M-parameter ConvNeXt-RS encoder improves
over a non-pretrained baseline by $+4.7$, $+5.0$, $+5.8$, and
$+1.4$\,pp on EuroSAT-S1, BigEarthNet-S1, DFC2020-S1, and Flood-S1
respectively, and outperforms the Copernicus-FM ViT (139M parameters,
frozen) on three of four tasks at roughly $5{\times}$ fewer
parameters. A pretraining data-scaling study reveals task-dependent
data efficiency: classification saturates with as little as 10--25\%
of the SSL4EO-S12 v1.1 corpus, while dense prediction continues to
improve significantly up to 25--50\%.
SparK-SAR's novelty therefore
lies not only in adapting sparse masked modeling to two-channel SAR
imagery through the ConvNeXt-RS encoder, but also in the extensive
empirical analysis accompanying it, spanning pretraining data
scaling, representation dynamics, and geographic data leakage.
\end{abstract}


\end{frontmatter}

\section{Introduction}
\label{sec:intro}
Unlike optical sensors, synthetic aperture radar (SAR) 
satellites image the land surface using
microwave backscatter rather than reflected sunlight, enabling
all-weather, day-and-night acquisition well suited to land-cover
mapping, flood assessment, and environmental change monitoring,
particularly in regions where persistent cloud cover limits optical
observation~\citep{moreira2013tutorial}. Since its 2014 launch, ESA's
Sentinel-1 mission has revisited most of the globe every 6--12 days,
accumulating a large and continually growing archive of freely
available Sentinel-1 Ground Range Detected (S1GRD)
imagery~\citep{esa2012sentinel}. This unlabeled archive is abundant,
but SAR imagery is difficult for non-experts to interpret visually,
making manual annotation of large SAR scenes highly labor intensive.
Publicly available labeled SAR benchmarks have therefore stayed small
relative to computer vision norms, which in turn limits how well
supervised deep learning models can perform on SAR downstream
tasks~\citep{wang2022self}.

Self-supervised learning (SSL) offers one route to transferable
representations without relying on manually labeled data, and masked image
modeling (MIM)~\citep{he2022mae,xie2022simmim} is a natural fit for
S1GRD pretraining specifically: it requires no S1GRD-appropriate
photometric augmentation scheme of the kind contrastive methods
depend on~\citep{wang2022self}, and it naturally yields the dense,
patch-level representations that earth observation's dense prediction
tasks need. Existing MIM implementations, however, are almost all
built on Vision Transformer (ViT)~\citep{dosovitskiy2021an} encoders, 
which typically demand large pretraining corpora and heavy compute to 
perform well; Copernicus-FM~\citep{wang2025copernicus}, for instance, 
needed 512 GPU hours on 4$\times$A100 GPUs and multi-modal inputs spanning 
four Sentinel missions (Sentinel-1, 2, 3, 5P) to reach competitive results.

For many research groups and applied teams working with SAR data,
matching the compute scale needed to pretrain these large foundation
models is itself out of reach. We therefore restrict this study's
primary comparisons to end-to-end finetuned convolutional neural
network (CNN) encoders rather than reporting frozen ViT foundation
model backbones as directly finetuned baselines. This is not solely a
matter of compute: Copernicus-Bench's own published learning-rate
search for these models ($10^{-4}$, $10^{-3}$, $10^{-2}$) was
validated only for frozen-backbone and decoder-head finetuning, not
for full end-to-end finetuning, so no full-finetuning result for
these architectures could be defended as using a properly validated
learning rate. Reporting one without a fair hyperparameter search
would conflate architectural differences with an unvalidated
finetuning setup rather than isolate them. A rigorous evaluation of
full fine-tuning for these foundation models, encompassing an
architecture-specific learning-rate search, appropriate regularization
schedules, and comparable per-model tuning budgets, would constitute a
dedicated study in its own right and lies outside this paper's scope.
The frozen-encoder results we do report (Table~\ref{tab:frozen}) are
included only as a reference point, not as the primary comparison.

CNN architectures, in contrast, bring stronger local inductive biases,
remain data efficient under modest compute budgets, and can be
finetuned end-to-end even on the small downstream datasets typical of
SAR applications. SparK~\citep{tian2023spark} was developed and
validated on natural-image benchmarks such as ImageNet~\citep{deng2009imagenet}, 
and porting its default UNet-768~\citep{ronneberger2015unet} 
decoder directly to S1GRD data at the prescribed
learning rate proves numerically unstable: FP16 mixed precision
training produces NaN losses, forcing either a redesigned decoder or
a different learning rate.

Building on these two lines of work, we introduce SparK-SAR: a
high-resolution sparse masked pretraining framework for S1GRD imagery
that combines SparK~\citep{tian2023spark} and
ConvNeXt~\citep{liu2022convnext} and is designed to run under a
constrained compute budget. We structure our contributions around
three questions: whether sparse convolutional masked pretraining,
previously developed and validated only on natural images, can be
adapted to produce transferable representations for S1GRD SAR imagery
under constrained compute; how much pretraining data and downstream
labeled data this benefit actually requires; and whether the observed
gains are genuine, rather than an artifact of geographic overlap
between the pretraining corpus and the downstream evaluation sets.

\textbf{Adapting sparse masked pretraining to S1GRD SAR imagery.}
\begin{itemize}
  \item \textbf{ConvNeXt-RS encoder}: a domain-adapted ConvNeXt
    encoder for SAR dense prediction, reducing stem stride from
    $4{\times}$ to $2{\times}$ to produce $14{\times}14$ feature
    maps (stride $16{\times}$) instead of the standard $7{\times}7$
    (stride $32{\times}$), providing $4{\times}$ more spatial tokens
    at identical parameter count~(Section~\ref{sec:encoder}).

  \item \textbf{Lightweight reconstruction decoders}: documentation of
    FP16 mixed precision pretraining instability with the standard UNet-768
    decoder on S1GRD data, and evaluation of UNet-128 and ConvNeXt-512 alternatives
    that stabilize pretraining, with ConvNeXt-512 reducing pretraining VRAM
    by 7.5--8.5\% per GPU at matched accuracy~(Section~\ref{sec:decoder}).

  \item \textbf{LLRD finetuning recipe}: a finding that layer-wise
    learning rate decay (LLRD) improves downstream task performance and
    reduces across-seed variance when finetuning the UNet-128-pretrained
    SparK encoder, but does not help a non-pretrained one~(Section~\ref{sec:finetune}).
\end{itemize}
Evaluated across four Copernicus-Bench S1GRD tasks~\citep{wang2025copernicus}
spanning classification, multi-label classification, semantic
segmentation, and change detection, the resulting 27.8M-parameter
encoder improves over a non-pretrained baseline by $+4.7$\,pp,
$+5.0$\,pp, $+5.8$\,pp, and $+1.4$\,pp on EuroSAT-S1, BigEarthNet-S1,
DFC2020-S1, and Flood-S1 respectively~(Section~\ref{sec:main}).
Despite using a 27.8M-parameter encoder, roughly \texorpdfstring{$5{\times}$}{5x}
smaller than Copernicus-FM ViT's 139M (frozen) parameters, our
finetuned model beats it on three of the four tasks, and it likewise
beats SoftCon~\citep{wang2024softcon} on three of four tasks and both
CROMA~\citep{fuller2023croma} and DOFA~\citep{xiong2024neural} on all
four. The comparatively modest Flood-S1 gain is examined further in a
dedicated ceiling ablation~(Section~\ref{sec:ablation_flood}), which
points to a boundary condition for this approach: single-image masked
pretraining appears to transfer less effectively to temporal change
detection than to single-scene interpretation tasks.

\textbf{Characterizing how much pretraining data and downstream supervision this benefit actually requires.}
\begin{itemize}
  \item \textbf{Pretraining data scaling study}: a systematic
    evaluation of pretraining data efficiency across eight data
    fractions (0--100\% of SSL4EO-S12 v1.1~\citep{blumenstiel2025ssl4eov11}) 
    with formal statistical analysis, revealing task dependent saturation behavior 
    in SAR amplitude representation learning~(Section~\ref{sec:scaling}).

  \item \textbf{Representation analysis}: Centered Kernel Alignment
    (CKA)~\citep{kornblith2019similarity} analysis across all data fractions, 
    characterizing how encoder representations evolve with pretraining data
    quantity~(Section~\ref{sec:cka}).

  \item \textbf{Label efficiency analysis}: a demonstration that SparK-SAR
  pretraining provides statistically significant accuracy gains over random
  initialization at every label fraction tested (1--50\%), with pretrained
  models matching full-label random-initialization performance using 10--50\%
  of labeled training data across all four
  tasks~(Section~\ref{sec:label_efficiency}).
\end{itemize}

\textbf{Verifying that the observed gains reflect genuine transfer rather than geographic leakage.}
\begin{itemize}
  \item \textbf{Geographic de-collision experiment}: identification of
    geographic overlap between the SSL4EO-S12 v1.1 pretraining corpus
    and the Copernicus-Bench evaluation sets, followed by
    re-pretraining SparK-SAR from scratch on a de-collided corpus with
    all overlapping tiles removed; the resulting change in downstream
    accuracy is below 0.1 percentage points on every task, confirming
    that the reported gains are not attributable to geographic data
    leakage~(Section~\ref{sec:leakage}).
\end{itemize}

A preliminary version of this work will be presented at SPIE Asia-Pacific
Remote Sensing 2026~\citep{natsume2026sparksarspie}. This journal 
submission substantially extends that conference paper with: 
(i) a systematic pretraining data scaling study across eight pretraining data 
fractions with formal statistical testing~(Section~\ref{sec:scaling}); 
(ii) a Centered Kernel Alignment representation analysis characterizing 
how encoder representations evolve with pretraining data 
quantity~(Section~\ref{sec:cka}); 
(iii) a full geographic de-collision analysis re-pretraining SparK-SAR
from scratch on a de-collided corpus with all overlapping pretraining
tiles removed, confirming that reported gains are not attributable to
geographic data leakage~(Section~\ref{sec:leakage});
(iv) a finetuning label efficiency analysis demonstrating that SparK-SAR pretraining 
provides statistically significant accuracy gains over random initialization 
at every label fraction and every task~(Section~\ref{sec:label_efficiency}); 
(v) a dedicated Flood-S1 ceiling ablation across loss functions and fusion 
strategies~(Section~\ref{sec:ablation_flood}); 
(vi) an expanded negative-results discussion; 
and (vii) extended methodological detail throughout.

\section{Related Work}
\label{sec:related}
\subsection{Masked Image Modeling}
Building on masked language modeling~\citep{devlin2019bert}, masked
image modeling (MIM) has become the leading approach to
self-supervised visual pretraining. MAE~\citep{he2022mae} showed that
reconstructing raw pixels from a small visible subset (as little as
25\%) of ViT patches produces representations that transfer well at
scale, and SimMIM~\citep{xie2022simmim} carried the same
reconstruction objective over to hierarchical Swin 
Transformers~\citep{liu2021swin}. I-JEPA~\citep{assran2023ijepa} 
took a different route, predicting
latent representations of the masked regions instead of pixel values,
which pushes the model toward semantic structure over low-level
texture.

These ViT-based approaches are effective but data- and
compute-hungry. Convolutional alternatives have narrowed that gap:
MCMAE~\citep{gao2022mcmae} combines convolution and transformer blocks
in a hybrid masked autoencoder, while SparK~\citep{tian2023spark}, the
framework this work builds on, shows that sparse masked modeling
works directly on purely hierarchical CNNs
(ResNet~\citep{he2016resnet}, ConvNeXt~\citep{liu2022convnext}) with no
transformer component at all. MMEarth~\citep{nedungadi2024mmearth} is
a concurrent example in the electro-optical remote sensing domain, pairing a
ConvNeXt V2~\citep{woo2023convnextv2} backbone with multi-modal MIM
objectives on electro-optical Sentinel-2 imagery; we pursue the same 
CNN MIM direction on a different modality, applying it to 
SAR-only pretraining under the SparK framework.

\subsection{Self-Supervised Learning for Remote Sensing}
In their survey of SSL for remote sensing, 
\citet{wang2022self} group existing methods into two dominant
paradigms, contrastive and generative. 

Among the contrastive methods,
SeCo~\citep{manas2021seco} builds positive pairs from seasonal
multi-temporal imagery, and GASSL~\citep{ayush2021gassl} instead treats
geographic co-location as a proxy for scene similarity. Both,
however, rely on augmentation pipelines built around electro-optical
statistics, which do not transfer cleanly to SAR backscatter.

MIM has also been adopted for electro-optical remote sensing.
SatMAE~\citep{cong2022satmae} adapted MAE for multi-spectral and
multi-temporal satellite data. Scale-MAE~\citep{reed2023scale}
introduced scale-aware positional encodings for images at varying
ground sample distances. FG-MAE~\citep{wang2025feature} proposed
feature-guided reconstruction targets better aligned with semantic
downstream tasks.

A separate line of work pairs Sentinel-1 SAR with electro-optical
data rather than treating SAR in isolation. SoftCon~\citep{wang2024softcon}
pretrains contrastively across multispectral and SAR imagery using
multi-label land-cover supervision as guidance. DOFA~\citep{xiong2024neural}
takes a different route, using a dynamic spectral adapter, so that a single
ViT can ingest whatever sensor modality it is given.
CROMA~\citep{fuller2023croma} instead unifies contrastive and
masked autoencoding objectives within one radar-optical model, and
Copernicus-FM~\citep{wang2025copernicus} pushes this further still,
training a multi-modal ViT across four Sentinel missions on 18.7M
aligned images at a reported cost of roughly 512 GPU hours on NVIDIA
A100 GPUs. These models perform well, but their multi-modal inputs
and near-exclusive reliance on frozen-encoder evaluation make them a
poor match for the SAR-only, constrained-compute setting this work
targets.

\subsection{SAR-Specific Self-Supervised Learning}
Compared to optical imagery, SSL research targeting SAR data
exclusively is still comparatively unexplored. 
\citet{allen2023large} demonstrated that large-scale MAE
pretraining on Sentinel-1 S1GRD data can substantially reduce label
requirements across two downstream tasks, vegetation-cover prediction
and land-cover classification, with the pretraining benefit
generalizing geographically and increasing, rather than merely
holding up, on regions outside the pretraining set.

SAR-JEPA~\citep{li2024predicting} introduced a joint-embedding 
predictive architecture (JEPA)-based approach for
SAR automatic target recognition (ATR) by predicting multi-scale
gradient features rather than raw pixels. SARATR-X~\citep{li2025saratrx}
subsequently incorporated multi-scale gradient features into a masked
image modeling framework, using a two-step self-supervised procedure
that first performed MIM pretraining on ImageNet to obtain a
favorable weight initialization, then continued MIM pretraining on a
curated corpus of unlabeled SAR target samples using the multi-scale
gradient features to suppress speckle noise, rather than the
joint-embedding predictive approach SAR-JEPA uses. However, both
models are trained and evaluated exclusively on ATR tasks, limiting
the generalizability to land-cover and change detection problems.

Applying DINO~\citep{caron2021dino}-style self-distillation to SAR
imagery, \citet{gallego2023exploring} and
\citet{martinez2023exploring} 
found only modest gains over training from scratch, 
with the resulting self-distilled embeddings
generalizing poorly to geographically unfamiliar regions without
further domain-specific adaptation.

To the best of our knowledge, SparK-SAR is among the first methods to
adapt sparse convolutional masked pretraining specifically for S1GRD
data and demonstrate its transferability across diverse downstream
SAR tasks under constrained compute.

\section{Materials and Methods}
\label{sec:methodology}
\begin{figure*}[ht]
\centering
\includegraphics[width=\linewidth]{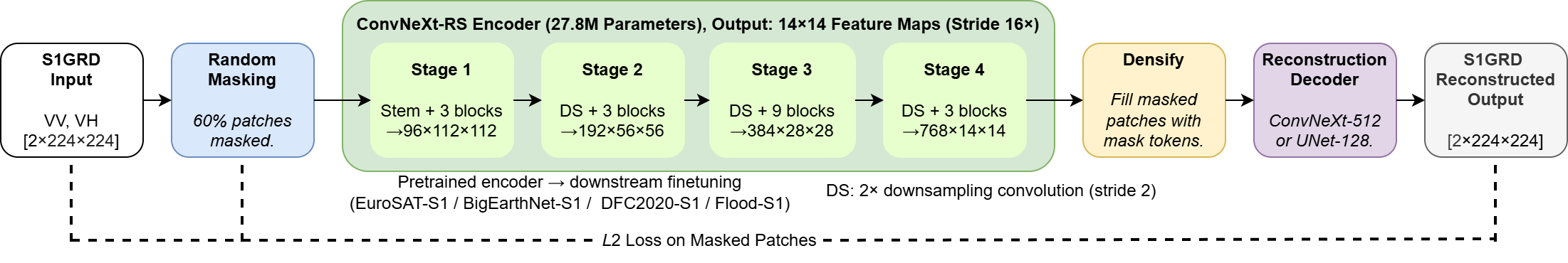}
\caption{The proposed SparK-SAR pretraining framework. The ConvNeXt-RS
encoder produces $14{\times}14$ feature maps (stride $16{\times}$) from
$224{\times}224$ S1GRD inputs. Only visible (unmasked) patches are
processed by the encoder via sparse convolutions; the decoder
reconstructs the full image including masked regions.}
\label{fig:framework}
\end{figure*}

Figure~\ref{fig:framework} illustrates how SparK-SAR adapts SparK's
sparse masked pretraining scheme~\citep{tian2023spark} to S1GRD data
under constrained compute. A $224{\times}224$ two-channel input tile, 
comprising VV (vertical transmit–vertical receive) and VH 
(vertical transmit–horizontal receive) backscatter in dB scale, 
is patch-masked at random, and the encoder
processes only the visible patches through sparse convolutions rather
than the full grid. The output feature maps are passed to a
reconstruction decoder tasked with predicting the masked patches.
Once pretraining finishes, the decoder is discarded and only the 
encoder is used for downstream finetuning.

\subsection{ConvNeXt-RS Encoder}
\label{sec:encoder}
ConvNeXt-Tiny's~\citep{liu2022convnext} default stem downsamples by
$4{\times}4$ before three further $2{\times}2$ stages, taking a
$224$-pixel input through $56{\to}28{\to}14{\to}7$ down to a final
$7{\times}7$ feature map at overall stride $32{\times}$; every output
token therefore corresponds to a $32{\times}32$-pixel patch of the
input, coarse enough to erase spatial detail that dense SAR tasks
such as segmentation and change detection depend on.

ConvNeXt-RS keeps everything about this backbone fixed except the
stem: halving its stride to $2{\times}2$ changes the spatial
trajectory to
\begin{equation}
  224 \to 112 \to 56 \to 28 \to 14 \quad (\text{stride}~16{\times})
  \label{eq:convnextrs}
\end{equation}
so the final feature map doubles in each spatial dimension to
$14{\times}14$, four times as many tokens as the stock $7{\times}7$
output, each now covering a finer $16{\times}16$-pixel patch instead
of $32{\times}32$; the overall stride drops correspondingly from
$32{\times}$ to $16{\times}$. Depth, per-stage block counts, and
channel widths are all untouched, so the parameter count stays at
27.8M, and the change is fully compatible with SparK's sparse
convolution mechanism operating on visible patches at every stage.

\begin{figure*}[ht]
\centering
\includegraphics[width=\linewidth]{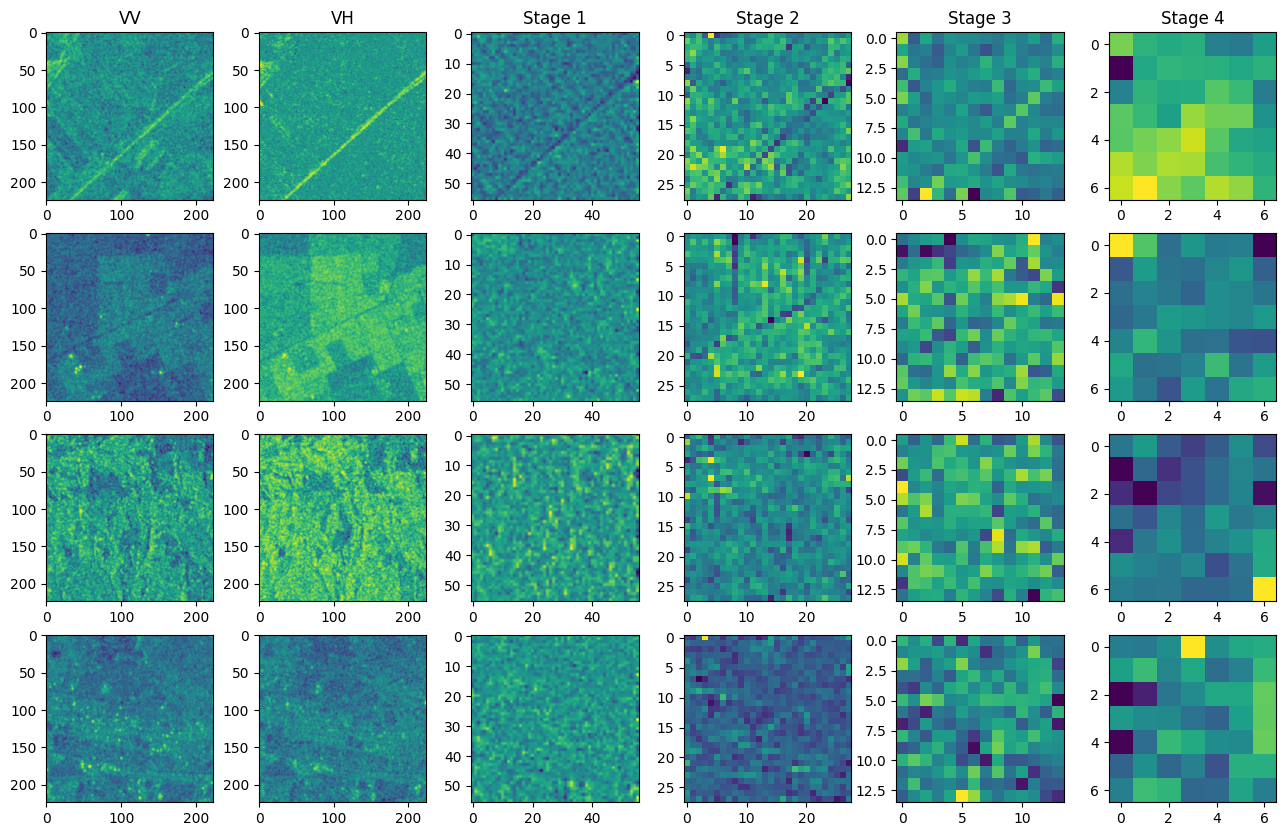}
\caption{Feature maps of the standard ConvNeXt-Tiny encoder ($7{\times}7$,
stride $32{\times}$). Columns from left to right: VV input, VH input,
and Stage 1--4 features. Stage 4 produces $7{\times}7$ maps that lose
spatial structure.}
\label{fig:features_7x7}
\end{figure*}

\begin{figure*}[ht]
\centering
\includegraphics[width=\linewidth]{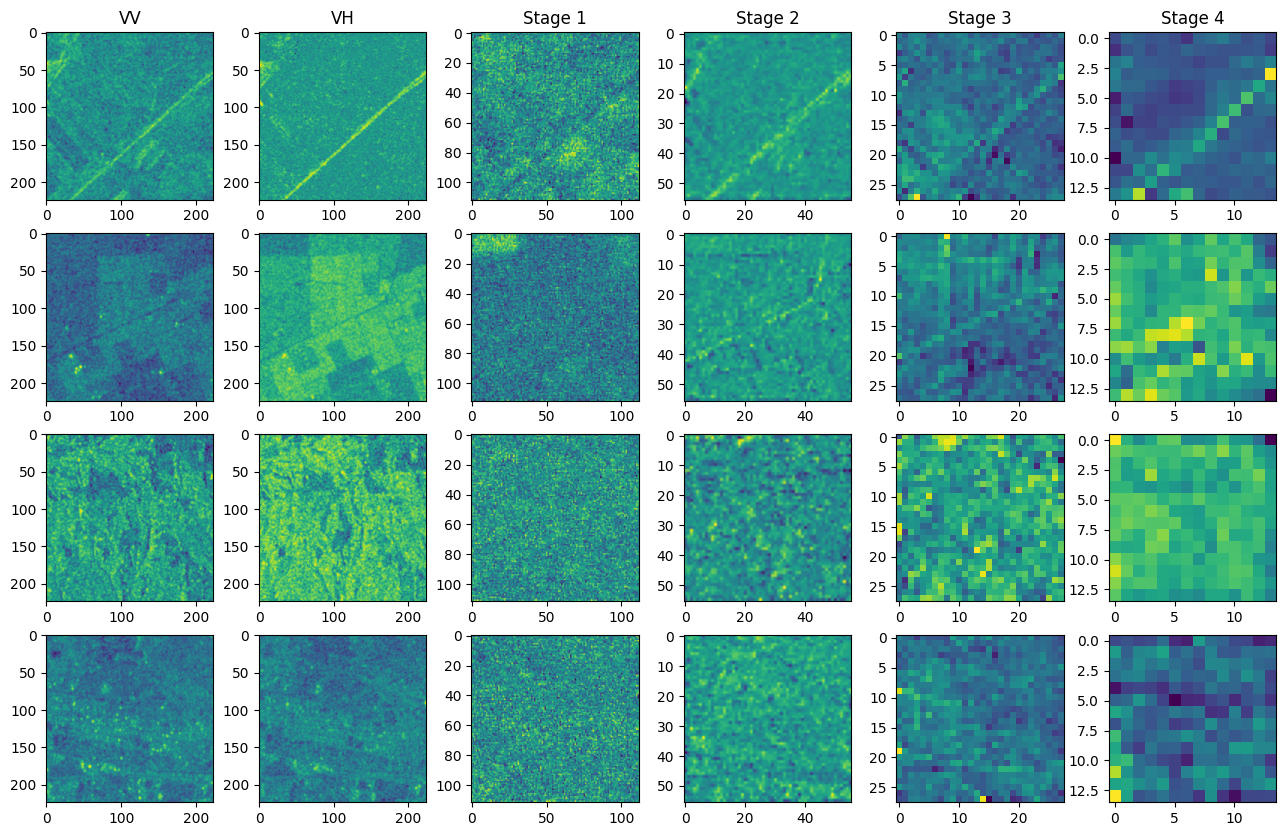}
\caption{Feature maps of the proposed ConvNeXt-RS encoder ($14{\times}14$,
stride $16{\times}$). Stage 4 retains recognizable spatial structure
corresponding to scene content, in contrast to Figure~\ref{fig:features_7x7}.}
\label{fig:features_14x14}
\end{figure*}

Figures~\ref{fig:features_7x7} and~\ref{fig:features_14x14} contrast
the resulting stage-by-stage feature maps for the two encoders:
ConvNeXt-Tiny's Stage 4 output has largely lost spatial structure by
$7{\times}7$, whereas ConvNeXt-RS's $14{\times}14$ Stage 4 output
still shows recognizable scene structure.

\begin{figure*}[ht]
\centering
\includegraphics[width=\linewidth]{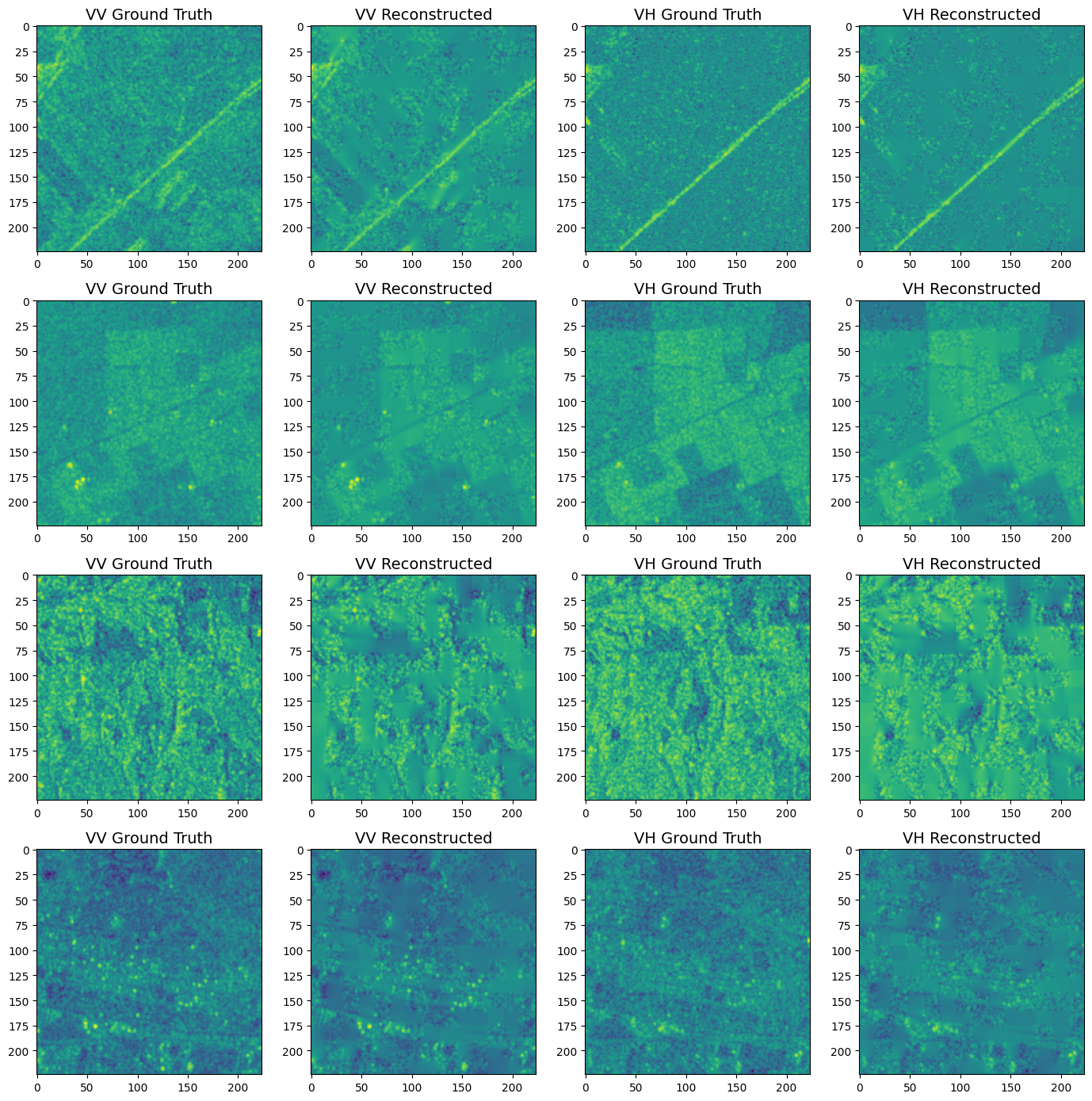}
\caption{SparK-SAR reconstruction examples on SSL4EO-S12 v1.1 S1GRD
patches. Each row shows (left to right) VV ground truth, VV
reconstructed, VH ground truth, and VH reconstructed (60\% masking
ratio). The reconstructions recover large-scale backscatter structure while
appearing smooth relative to ground truth, consistent with the L2
reconstruction objective of MAE-family methods~\citep{he2022mae}.}
\label{fig:reconstruction}
\end{figure*}

\subsection{Reconstruction Decoders}
\label{sec:decoder}
SparK-SAR supports interchangeable reconstruction decoders.
We evaluate three designs.

\textbf{UNet-768}, SparK's original decoder, is a hierarchical 
UNet~\citep{ronneberger2015unet} that aggregates multi-scale encoder features 
through 768 channels. 
Under the optimal learning rate $lr{=}10^{-3}$, which is of the same
order as that used by the original SparK configuration~\citep{tian2023spark},
UNet-768 encounters gradient overflow (NaN loss) at epoch 5--12 under FP16 
mixed-precision training, coinciding with the linear-warmup-to-cosine-decay 
schedule remaining near its peak learning rate in the epochs immediately 
following the 4-epoch warmup. We did not log gradient norms directly, 
but the timing is consistent with gradient magnitudes in the 
high-dimensional skip connections exceeding the FP16 dynamic range
(${\leq}65{,}504$).

At the reduced pretraining learning rate $lr{=}10^{-4}$, UNet-768 trains stably
under FP16 mixed precision pretraining,
but yields inferior downstream performance compared to
UNet-128 at the same learning rate (Table~\ref{tab:ablation_lr}),
confirming that decoder capacity itself, not solely the lower learning rate
required for stability, contributes to UNet-768's inferior downstream
performance.

To characterize its VRAM requirements beyond our production hardware's
32\,GB ceiling, we profiled UNet-768 on two A100 80\,GB GPUs
(Table~\ref{tab:vram}). Running in FP32 (80,599\,MiB/GPU) or BF16
(45,091\,MiB/GPU) on this hardware eliminates the instability, but at
VRAM costs that would still exceed our production 4$\times$V100\,32\,GB
budget.
The near-identical VRAM consumption of UNet-768 under FP16
(45,037\,MiB) and BF16 (45,091\,MiB) confirms that the FP16
instability is a numerical precision issue rather than a memory
constraint.
We therefore adopt UNet-128 and ConvNeXt-512, which remain stable
under FP16 at $lr{=}10^{-3}$ and achieve superior downstream
performance.

\textbf{UNet-128} keeps the same hierarchical, multi-scale
aggregation as UNet-768 but shrinks the channel width to 128; that
reduction in decoder capacity is enough on its own to keep FP16
pretraining stable at $lr{=}10^{-3}$.

\textbf{ConvNeXt-512} is our proposed non-hierarchical decoder
comprising a single ConvNeXt block~\citep{liu2022convnext} with 512 channels 
followed by PixelShuffle upsampling (upscale factor $16{\times}$ for 
$14{\times}14$ encoder features reconstructing $224{\times}224$ inputs). 
Because the ConvNeXt-512 decoder works from only the encoder's final feature 
map rather  than aggregating across scales, ConvNeXt-512 pretrains at 
7.5--8.5\% lower VRAM per GPU than UNet-128 (the exact figure depends on hardware)
without sacrificing downstream accuracy. Table~\ref{tab:vram} reports
VRAM consumption for both decoder configurations across hardware.

\begin{table}[t]
\caption{
Pretraining VRAM usage per GPU for different decoder configurations, 
hardware, and numerical precision. 
Our production setup consists of 4$\times$V100\,32\,GB GPUs.
The global batch size is 256 samples across all GPUs. 
}
\label{tab:vram}
\centering
\scriptsize
\setlength{\tabcolsep}{2pt}
\begin{tabularx}{\columnwidth}{@{}l X c r l@{}}
\toprule
\textbf{Decoder} & \textbf{Hardware} & \textbf{Prec.}
& \textbf{VRAM/GPU} & \textbf{Status} \\
\midrule
UNet-768 & A100 80\,GB $b=128$ & FP16
& 45,037\,MiB & NaN (ep.\ 5--12) \\
UNet-768 & A100 80\,GB $b=128$ & BF16
& 45,091\,MiB & Stable \\
UNet-768 & A100 80\,GB $b=128$ & FP32
& 80,599\,MiB & Stable \\
\midrule
UNet-128 & V100 32\,GB, $b=64$ & FP16
& 17,460\,MiB & Stable \\
ConvNeXt-512 & V100 32\,GB, $b=64$ & FP16
& 15,974\,MiB & Stable \\
\textit{Saving} & & &
\textit{1,486\,MiB (8.5\%)} & \\
\midrule
UNet-128 & A100 80\,GB, $b=256$ & FP16
& 62,617\,MiB & Stable \\
ConvNeXt-512 & A100 80\,GB, $b=256$ & FP16
& 57,909\,MiB & Stable \\
\textit{Saving} & & &
\textit{4,708\,MiB (7.5\%)} & \\
\bottomrule
\end{tabularx}
\end{table}

\subsection{Pretraining Setup}
\label{sec:pretrain}
Pretraining data comes from the SSL4EO-S12 v1.1 S1GRD
subset~\citep{blumenstiel2025ssl4eov11}: 243,968 globally distributed
locations, each with four seasonal (VV, VH) backscatter acquisitions at
$264{\times}264$ pixels in decibel (dB) scale. Rather than using all four
acquisitions per location every iteration, we draw one season
uniformly at random each time, which gives the model implicit
temporal augmentation for free, without needing paired imagery. On
top of that we apply per-channel normalization (using the empirically
measured statistics $\mu_{\mathrm{VV}}{=}{-}12.577$,
$\sigma_{\mathrm{VV}}{=}5.179$, $\mu_{\mathrm{VH}}{=}{-}20.265$,
$\sigma_{\mathrm{VH}}{=}5.872$), a random resized crop down to
$224{\times}224$, and random horizontal and vertical flips.

Optimization uses AdamW with $lr{=}10^{-3}$, weight decay $0.04$, and
$\beta{=}[0.9,0.999]$; a global batch size of 256; 60\% masking; an
$L_2$ reconstruction objective; gradient clipping at 5.0; and FP16
mixed precision, run for 50 epochs (4 of them warmup) under PyTorch
2.9.1 with CUDA 12.6.

Table~\ref{tab:recipe} contrasts these choices with the original SparK
ImageNet configuration.
Figure~\ref{fig:reconstruction} shows representative reconstruction 
examples from the pretrained SparK-SAR model on SSL4EO-S12 v1.1 
S1GRD patches.

Pretraining is comparatively inexpensive. The UNet-128 and 
ConvNeXt-512 configurations complete in 8.9 and 7.9 wall clock hours 
respectively on 4$\times$ NVIDIA V100 (32~GB) GPUs 
(35.6 and 31.6 GPU-hours; Table~\ref{tab:compute}), well below the 
8$\times$A100 (80~GB) multi-node configurations reported for 
CROMA~\citep{fuller2023croma} (300--600 epochs) and 
SoftCon~\citep{wang2024softcon} (100 epochs), though neither paper reports
absolute GPU-hour or wall clock figures, precluding an exact comparison.

\begin{table}[ht]
\caption{SparK-SAR pretraining recipe compared with the original SparK
ImageNet configuration~\citep{tian2023spark}.}
\label{tab:recipe}
\centering
\begin{tabular}{lcc}
\toprule
& \textbf{SparK (ImageNet)} & \textbf{SparK-SAR (S1GRD)} \\
\midrule
Dataset size    & 1.28M images        & 243,968 locations \\
Epochs          & 1,600               & 50 \\
Batch size      & 4,096               & 256 \\
Optimizer       & LAMB                & AdamW \\
Max LR          & $3.2{\times}10^{-3}$ & $10^{-3}$ \\
Encoder output  & $7{\times}7$        & $14{\times}14$ \\
Decoder         & UNet-768            & UNet-128 / ConvNeXt-512 \\
Precision       & FP16                & FP16 \\
Hardware        & Unreported          & 4$\times$V100 32\,GB \\
\bottomrule
\end{tabular}
\end{table}

\begin{table}[t]
\centering
\caption{Pretraining compute cost. Both configurations pretrain on
4$\times$ NVIDIA V100 (32~GB) GPUs.}
\label{tab:compute}
\begin{tabular}{lcc}
\toprule
Decoder & Wall-clock (h) & GPU-hours \\
\midrule
UNet-128 & 8.9 & 35.6 \\
ConvNeXt-512 & 7.9 & 31.6 \\
\bottomrule
\end{tabular}
\end{table}

\subsection{Finetuning Protocol}
\label{sec:finetune}
Every pretrained ConvNeXt-RS encoder is then finetuned end-to-end,
separately, on each downstream task: AdamW at $lr{=}10^{-3}$ (weight
decay $10^{-4}$), OneCycleLR with final\_div\_factor $=100$, 50
epochs (4 warmup), FP32, under a separate codebase running PyTorch
2.1.2 and CUDA 11.8. Pretraining and finetuning therefore run on different framework
versions; we did not isolate the effect of this difference.

Task-specific finetuning follows the Copernicus-Bench protocol: EuroSAT-S1
uses a linear classification head, while BigEarthNet-S1's multi-label
classification task is trained with a multi-label soft-margin loss. 
The two dense prediction tasks, DFC2020-S1 and Flood-S1, 
both employ a UPerNet~\citep{xiao2018upernet}
decoder augmented with an auxiliary FCN head. For DFC2020-S1 this decoder
operates directly on the multi-scale features produced by the ConvNeXt-RS
encoder; for Flood-S1, it instead operates on the difference between
multi-scale features computed separately for the pre- and post-event
images. The ConvNeXt-RS backbone is finetuned end-to-end throughout
instead of being kept frozen.

Inputs are resized to the resolution specified by the Copernicus-Bench
benchmark protocol for each task. EuroSAT-S1 (native $64{\times}64$) and
BigEarthNet-S1 (native $120{\times}120$) are upsampled to $224{\times}224$
using bilinear interpolation. Flood-S1 is already native at
$224{\times}224$, so no resizing is applied. 
DFC2020-S1 is kept at its native $256{\times}256$ resolution rather than
being unified to $224{\times}224$ like the other three tasks.
We retain each task's benchmark-specified resolution, 
rather than unifying all four to a single value, 
to keep our evaluation as directly comparable to the published
Copernicus-Bench baselines as our full-finetuning setting allows
(Section~\ref{sec:intro}). For DFC2020-S1 and Flood-S1, resizing (where
applied) is performed jointly on the image and the segmentation mask,
using paired image-mask transformations, with bilinear interpolation for
the image and nearest-neighbor interpolation for the mask so that resizing
does not blend adjacent class labels into fractional, non-existent
intermediate values at boundaries. Because ConvNeXt-RS is a fully
convolutional encoder with no fixed-input-size constraint from positional
embeddings, it accepts both resolutions directly, with the stride-16
feature map scaling accordingly ($14{\times}14$ for $224{\times}224$
inputs vs.\ $16{\times}16$ for DFC2020-S1's $256{\times}256$).

We apply Layer-wise Learning Rate Decay
(LLRD)~\citep{bao2022beit,clark2020electra} with layer decay factor
$0.75$, scaling the learning rate geometrically from the classification
head toward the stem.
LLRD runs also exclude the backbone's biases and normalization parameters
from weight decay. For the UNet-128-pretrained encoder, LLRD improves mean
accuracy on all four tasks and reduces across-seed variance, most markedly
on DFC2020-S1 and BigEarthNet-S1 (Section~\ref{sec:ablation_llrd}); it does
not improve a non-pretrained encoder (Table~\ref{tab:main} and the 0\% row of
Table~\ref{tab:scaling}) and lowers its Flood-S1 mIoU by 0.8\,pp, so we treat
it as a finetuning recipe for pretrained encoders.
All results are reported as mean$\,\pm\,$sample standard deviation ($n{-}1$ denominator) over 5 fixed random seeds
(42, 123, 456, 789, 1337), with deterministic seed initialization for 
PyTorch, CUDA, NumPy, Python’s random, data-loader workers, 
and the distributed data sampler.

Our production results were obtained on V100 GPUs (4$\times$V100 32\,GB for
pretraining, a single V100 for finetuning). After these GPUs were retired,
the Flood-S1 ablation (Table~\ref{tab:ablation_flood}), the LLRD
decay-factor sweep (Table~\ref{tab:llrd_lambda_sweep}, all columns except
the production $\lambda{=}0.75$), the label-efficiency experiments
(Table~\ref{tab:labeleff}, all rows except the 100\% row, which reuses the
production runs) and the frozen-encoder evaluation
(Table~\ref{tab:frozen}) were run on NVIDIA RTX A6000 48\,GB GPUs with the
same precision settings (FP16 pretraining, FP32 finetuning).
The de-collision experiment (Section~\ref{sec:leakage}) was pretrained on one
A100 80\,GB GPU and finetuned on A6000 GPUs.

\FloatBarrier
\section{Results}
\label{sec:experiments}
\subsection{Datasets}
\label{sec:datasets}
Table~\ref{tab:datasets} summarizes the four
Copernicus-Bench~\citep{wang2025copernicus} S1GRD tasks used for
evaluation, each built from its own established source dataset:
EuroSAT~\citep{helber2019eurosat} (EuroSAT-S1),
BigEarthNet~\citep{sumbul2019bigearthnet} (BigEarthNet-S1), the 2020
IEEE GRSS Data Fusion Contest dataset~\citep{robinson2021dfc2020}
(DFC2020-S1), and the Kuro Siwo flood mapping
dataset~\citep{bountos2024kurosiwo} (Flood-S1).

The validation split is used solely for checkpoint selection 
(best validation metric). No hyperparameters were tuned on 
the validation set after the pretraining configuration was fixed.

\begin{table}[t]
\caption{Copernicus-Bench S1GRD downstream evaluation tasks.}
\label{tab:datasets}
\centering
\scriptsize
\setlength{\tabcolsep}{2.5pt}
\begin{tabularx}{\columnwidth}{@{}l X l r r r c@{}}
\toprule
\textbf{Dataset} & \textbf{Task} & \textbf{Metric}
& \textbf{Train} & \textbf{Val} & \textbf{Test} & \textbf{Image size} \\
\midrule
EuroSAT-S1
& Classification
& OA (\%) & 16,200 & 5,400 & 5,400 & $64{\times}64$ \\
BigEarthNet-S1
& Multi-label classif.
& mAP (\%) & 11,894 & 6,117 & 5,991 & $120{\times}120$ \\
DFC2020-S1
& Segmentation
& mIoU (\%) & 3,156 & 986 & 986 & $256{\times}256$ \\
Flood-S1
& Change detection
& mIoU (\%) & 3,000 & 1,000 & 1,000 & $224{\times}224$ \\
\bottomrule
\end{tabularx}
\end{table}

\subsection{Baselines}
\label{sec:baselines}
We compare against:
(i)~a non-pretrained ConvNeXt-RS baseline finetuned with identical
protocol, isolating the pretraining contribution;
(ii)~a standard ConvNeXt ($7{\times}7$, stride $32{\times}$) pretrained
with SparK-SAR at matched learning rate ($lr{=}10^{-3}$), isolating the
resolution contribution;
(iii)~four published frozen-encoder results drawn directly from
their source papers: Copernicus-FM ViT-B/16 (139M
parameters)~\citep{wang2025copernicus}, SoftCon ViT-B/14
(86M)~\citep{wang2024softcon}, CROMA ViT-B/8
(86M)~\citep{fuller2023croma}, and DOFA ViT-B/16
(111M)~\citep{xiong2024neural}.

As ConvNeXt-RS is a novel domain-adapted architectural variant, no pre-existing
ImageNet-pretrained weights are available with the $2{\times}$ stem
stride and 2-channel (VV, VH) S1GRD input. 
The non-pretrained random initialization
baseline therefore serves as the appropriate control for isolating the
SAR-specific pretraining contribution.

\subsection{Main Results}
\label{sec:main}
Table~\ref{tab:main} presents our main results.
SparK-SAR with ConvNeXt-RS and LLRD finetuning improves over the
non-pretrained baseline by $+4.7$\,pp, $+5.0$\,pp, $+5.8$\,pp, and
$+1.4$\,pp on EuroSAT-S1, BigEarthNet-S1, DFC2020-S1, and Flood-S1
respectively (UNet-128 decoder).
The ConvNeXt-512 decoder achieves comparable performance while
reducing pretraining VRAM by 7.5--8.5\% per GPU, depending on hardware.

Despite having roughly $5{\times}$ fewer parameters than
Copernicus-FM ViT (27.8M vs.\ 139M, frozen encoder), our finetuned
model still comes out ahead on three of the four tasks: EuroSAT-S1
(\ms{88.49}{0.16} vs.\ $87.2{\pm}0.1$\%), DFC2020-S1
(\ms{56.00}{0.08} vs.\ $52.4{\pm}0.1$\%), and Flood-S1
(\ms{80.21}{0.10} vs.\ $77.7{\pm}0.0$\%).

On BigEarthNet-S1, SparK-SAR (\ms{77.03}{0.40}) falls marginally
below Copernicus-FM ($77.9{\pm}0.0$\%) and SoftCon ($78.7{\pm}0.0$\%)
under frozen evaluation.
We note that Copernicus-FM incorporates tile geographic coordinates
as an additional input modality via hypernetwork
encodings~\citep{wang2025copernicus}.
BigEarthNet-S1 labels are derived from the CORINE Land Cover
map~\citep{sumbul2019bigearthnet,eea2020corine}, which exhibits strong
geographic clustering of land-cover classes across Europe.
It is therefore possible that Copernicus-FM's frozen score on this
task partly reflects a geographic coordinate prior rather than purely
learned SAR representations.
SoftCon's strong result likely reflects its multi-label land-cover
guidance during pretraining~\citep{wang2024softcon}, which directly
benefits this multi-label classification task.
These hypotheses are not formally tested and are presented as caveats.

Table~\ref{tab:frozen} reports frozen encoder evaluation results for 
SparK-SAR. Under this protocol, ConvNeXt-RS trails frozen ViT baselines 
on classification tasks, consistent with the known disadvantage of 
convolutional architectures under linear probing~\citep{tian2023spark}. 
On dense prediction tasks, ConvNeXt-RS remains competitive with all 
frozen ViT baselines on DFC2020-S1 and exceeds all frozen ViT baselines 
on Flood-S1 ($78.38{\pm}0.12$ vs.\ $77.7{\pm}0.0$ for Copernicus-FM), 
demonstrating that SparK-SAR pretraining produces spatially rich 
representations even without task-specific adaptation.

Returning to the finetuned comparison, the same pattern holds against
the other three frozen ViT baselines, all of which have substantially
more parameters than our encoder: our finetuned model beats CROMA and
DOFA on every task, and beats SoftCon on three of the four (EuroSAT-S1,
DFC2020-S1, Flood-S1). BigEarthNet-S1 is the sole task where a frozen
baseline, SoftCon at $78.7{\pm}0.0$\%, comes out ahead, for the reasons
discussed above.

\begin{table*}[t]
\centering
\caption{Main results on Copernicus-Bench S1GRD tasks. Frozen baselines (top) evaluate a linear probe (classification) or UPerNet head (dense prediction) on frozen pretrained features; SparK-SAR (bottom) is finetuned end-to-end. Because these protocols are not directly comparable, best results are bolded independently within each block.}
\label{tab:main}
\resizebox{\textwidth}{!}{%
\begin{tabular}{lccccc}
\hline
\textbf{Method} & \textbf{Params} & \textbf{EuroSAT-S1} & \textbf{BigEarthNet-S1} & \textbf{DFC2020-S1} & \textbf{Flood-S1} \\
 & & OA (\%) & mAP (\%) & mIoU (\%) & mIoU (\%) \\
\hline
\multicolumn{6}{l}{\textit{Frozen baselines (linear probe / UPerNet head)}} \\
\hline
Copernicus-FM ViT-B/16~\citep{wang2025copernicus} & 139M & \textbf{87.2$\pm$0.1} & 77.9$\pm$0.0 & 52.4$\pm$0.1 & \textbf{77.7$\pm$0.0} \\
SoftCon ViT-B/14~\citep{wang2024softcon} & 86M & 83.6$\pm$0.1 & \textbf{78.7$\pm$0.0} & \textbf{52.8$\pm$0.6} & 77.2$\pm$0.1 \\
CROMA ViT-B/8~\citep{fuller2023croma} & 86M & 83.9$\pm$0.1 & 70.8$\pm$0.0 & 52.7$\pm$0.1 & 77.4$\pm$0.1 \\
DOFA ViT-B/16~\citep{xiong2024neural} & 111M$^\S$ & 81.7$\pm$0.1 & 70.5$\pm$0.0 & 49.7$\pm$0.1 & 76.0$\pm$0.1 \\
\hline
\multicolumn{6}{l}{\textit{Finetuned end-to-end (Ours)}} \\
\hline
No pretraining$^\dagger$ & 27.8M & 83.79$\pm$0.53 & 72.03$\pm$0.09 & 50.24$\pm$0.25 & 78.86$\pm$0.32 \\
SparK-SAR + UNet-128, LLRD & 27.8M & 88.49$\pm$0.16 & \textbf{77.03$\pm$0.40} & \textbf{56.00$\pm$0.08} & \textbf{80.21$\pm$0.10} \\
SparK-SAR + ConvNeXt-512, LLRD$^\ddagger$ & 27.8M & \textbf{88.63$\pm$0.34} & 76.60$\pm$0.37 & 55.94$\pm$0.08 & 80.11$\pm$0.10 \\
\hline
\end{tabular}%
}
\\[2pt]
{\footnotesize $^\S$DOFA parameter count from \citet{lu2025vision}; 
neither the original DOFA paper nor Copernicus-FM report it directly. 
$^\dagger$Finetuned without LLRD. 
$^\ddagger$Saves 7.5--8.5\% VRAM vs.\ UNet-128 (Table~\ref{tab:vram}), 
depending on hardware.}
\end{table*}

\begin{table*}[ht]
\caption{Frozen encoder evaluation of SparK-SAR ConvNeXt-RS 
(mean$\,\pm\,$std, 5 seeds). The backbone is frozen and only the 
task head is trained: a linear head for classification tasks and 
UPerHead~\citep{xiao2018upernet} for dense prediction tasks. 
Linear probing is known to disadvantage convolutional architectures 
relative to ViTs due to the absence of a global summary 
token~\citep{tian2023spark}; results should be interpreted alongside 
the full finetuning results in Table~\ref{tab:main}.}
\label{tab:frozen}
\centering
\small
\begin{tabular}{lcccc}
\toprule
\textbf{Config} & \textbf{EuroSAT-S1} & \textbf{BigEarthNet-S1}
  & \textbf{DFC2020-S1} & \textbf{Flood-S1} \\
& \textbf{OA (\%)} & \textbf{mAP (\%)}
  & \textbf{mIoU (\%)} & \textbf{mIoU (\%)} \\
\midrule
ConvNeXt-RS (frozen)
  & $76.56{\pm}0.09$ & $66.17{\pm}0.03$
  & $51.94{\pm}0.09$ & $\mathbf{78.38{\pm}0.12}$ \\
\midrule
\multicolumn{5}{l}{\textit{Frozen ViT baselines (from Table~\ref{tab:main})}} \\
Copernicus-FM ViT-B/16
  & $\mathbf{87.2{\pm}0.1}$ & $77.9{\pm}0.0$
  & $52.4{\pm}0.1$ & $77.7{\pm}0.0$ \\
SoftCon ViT-B/14
  & $83.6{\pm}0.1$ & $\mathbf{78.7{\pm}0.0}$
  & $\mathbf{52.8{\pm}0.6}$ & $77.2{\pm}0.1$ \\
CROMA ViT-B/8
  & $83.9{\pm}0.1$ & $70.8{\pm}0.0$
  & $52.7{\pm}0.1$ & $77.4{\pm}0.1$ \\
DOFA ViT-B/16
  & $81.7{\pm}0.1$ & $70.5{\pm}0.0$
  & $49.7{\pm}0.1$ & $76.0{\pm}0.1$ \\
\bottomrule
\end{tabular}
\end{table*}

\subsection{Ablation Studies}
\label{sec:ablation}
\subsubsection{
Output Resolution: \texorpdfstring{$7\times7$}{7x7} 
vs \texorpdfstring{$14\times14$}{14x14}
}
\label{sec:ablation_resolution}
Because Table~\ref{tab:ablation_lr} pretrains the standard $7{\times}7$
ConvNeXt encoder and our $14{\times}14$ ConvNeXt-RS encoder under
otherwise identical conditions (UNet-128 decoder, pretraining $lr{=}10^{-3}$), it
doubles as a resolution ablation. The higher-resolution encoder's edge
is concentrated in the dense prediction tasks ($+1.4$\,pp DFC2020-S1,
$+0.8$\,pp Flood-S1) and much smaller for classification ($+0.3$\,pp
EuroSAT-S1), which is what we would expect if the extra spatial
tokens mainly help pixel-level rather than scene-level prediction.

The matched learning rate control largely rules out the pretraining learning rate
as a confounding factor: the original $7{\times}7$ experiments used a pretraining
$lr{=}10^{-4}$ (required to stabilize training), while our
$14{\times}14$ ConvNeXt-RS uses a pretraining $lr{=}10^{-3}$ (all models are
finetuned at $lr{=}10^{-3}$).
Table~\ref{tab:ablation_lr} shows that pretraining the $7{\times}7$ encoder at
$lr{=}10^{-3}$ with UNet-128 still underperforms $14{\times}14$ at
the same pretraining learning rate, confirming that the resolution modification
is the primary contributor.
At a pretraining $lr{=}10^{-4}$, UNet-128 outperforms UNet-768 despite its lower 
decoder capacity, indicating that the larger decoder's deficit is not 
explained by its reduced learning rate alone.

\begin{table*}[ht]
\caption{Output resolution and decoder ablation at pretraining learning rates
$10^{-4}$ and $10^{-3}$; all encoders were finetuned at $lr{=}10^{-3}$ without LLRD
(mean$\,\pm\,$std, 5 seeds). 
All encoders were pretrained with FP16 mixed precision on
4$\times$V100 32\,GB GPUs and finetuned in FP32 on V100 GPUs.}
\label{tab:ablation_lr}
\centering
\small
\begin{tabular}{lcccc}
\toprule
\textbf{Config} & \textbf{EuroSAT-S1} & \textbf{BigEarthNet-S1}
  & \textbf{DFC2020-S1} & \textbf{Flood-S1} \\
\midrule
$7{\times}7$, pretrain $lr{=}10^{-4}$, UNet-768
  & $84.57{\pm}0.25$ & $72.27{\pm}0.45$ & $52.12{\pm}0.03$ & $78.67{\pm}0.16$ \\
$7{\times}7$, pretrain $lr{=}10^{-4}$, UNet-128
  & $85.51{\pm}0.11$ & $72.45{\pm}0.25$ & $52.74{\pm}0.08$ & $78.72{\pm}0.21$ \\
$7{\times}7$, pretrain $lr{=}10^{-3}$, UNet-128
  & $87.85{\pm}0.30$ & $74.89{\pm}0.38$ & $54.22{\pm}0.18$ & $79.33{\pm}0.18$ \\
$14{\times}14$, pretrain $lr{=}10^{-3}$, UNet-128
  & $\mathbf{88.11{\pm}0.29}$ & $\mathbf{75.47{\pm}0.70}$ & $\mathbf{55.59{\pm}0.23}$
  & $\mathbf{80.11{\pm}0.12}$ \\
\bottomrule
\end{tabular}
\end{table*}

\subsubsection{Decoder Design and Efficiency}
\label{sec:ablation_decoder}
UNet-128 and ConvNeXt-512 decoders achieve comparable downstream
accuracy across all four tasks (Table~\ref{tab:main}), despite
ConvNeXt-512 operating on only the final encoder stage rather than
aggregating multi-scale features. This result is consistent with the
hierarchical decoder not being necessary for strong downstream
accuracy: the lightweight ConvNeXt-512 decoder achieves statistically
indistinguishable performance (paired $t$-test, $p{=}0.17$) while
reducing VRAM by 7.5--8.5\% per GPU, depending on hardware
(Table~\ref{tab:vram}). We note that decoder capacity is not perfectly
disentangled from learning rate in this comparison, since the
higher-capacity UNet-768 decoder (Table~\ref{tab:ablation_lr}) required
a lower learning rate for FP16 stability; the finding above is
therefore best read as evidence against a decoder-capacity effect at
matched learning rate and precision (UNet-128 vs.\ ConvNeXt-512, both
at $lr{=}10^{-3}$), rather than a fully controlled ablation across the
entire decoder-capacity range.

\subsubsection{LLRD Finetuning}
\label{sec:ablation_llrd}
Table~\ref{tab:ablation_llrd} compares finetuning with and without LLRD
for ConvNeXt-RS pretrained with UNet-128.
LLRD improves mean performance in the same direction and reduces
result variance across all four tasks. This improvement
is statistically significant after Benjamini--Hochberg false discovery rate 
(BH FDR) correction~\citep{benjamini1995controlling}
(paired $t$-tests, 5 seeds, family of four) for BigEarthNet-S1
($t{=}6.22$, $p_{BH}{=}0.014$) and DFC2020-S1 ($t{=}3.88$,
$p_{BH}{=}0.036$), but not for EuroSAT-S1 ($t{=}2.91$,
$p_{BH}{=}0.058$; significant before correction at $p{=}0.044$) or
Flood-S1 ($t{=}1.26$, $p_{BH}{=}0.28$), consistent with the smaller
absolute gains observed on these two tasks.
This consistency holds specifically across the 5-seed UNet-128
evaluations in Table~\ref{tab:ablation_llrd}.
A parallel 5-seed comparison on the ConvNeXt-512 pretrained checkpoint
(Table~\ref{tab:ablation_flood}, Subtraction+CE) shows the same direction
($80.11\pm0.15$ with LLRD vs.\ $79.99\pm0.14$ without) but a smaller gap
that is not statistically significant (paired $t$-test, $p{=}0.21$); on this
checkpoint the Flood-S1 benefit of LLRD is therefore small and cannot be
distinguished from seed noise. Both arms were run on an NVIDIA RTX A6000 GPU
after our V100 GPUs were retired; the corresponding V100 production run in
Table~\ref{tab:main} gives $80.11\pm0.10$.

\begin{table}[t]
\caption{Effect of LLRD on ConvNeXt-RS finetuning (UNet-128 pretrained,
mean$\pm$std, 5 seeds).}
\label{tab:ablation_llrd}
\centering
\scriptsize
\setlength{\tabcolsep}{2pt}
\begin{tabular}{@{}lcccc@{}}
\toprule
\textbf{Config} & \textbf{EuroSAT-S1} & \textbf{BigEarthNet-S1}
  & \textbf{DFC2020-S1} & \textbf{Flood-S1} \\
\midrule
No LLRD
  & $88.11\pm0.29$ & $75.47\pm0.70$
  & $55.59\pm0.23$ & $80.11\pm0.12$ \\

\makecell[l]{LLRD\\($\lambda=0.75$)}
  & $\mathbf{88.49\pm0.16}$ & $\mathbf{77.03\pm0.40}$
  & $\mathbf{56.00\pm0.08}$ & $\mathbf{80.21\pm0.10}$ \\
\bottomrule
\end{tabular}
\end{table}

\begin{table}[t]
\centering
\caption{Sensitivity of downstream performance to the LLRD
decay factor $\lambda$, evaluated on DFC2020-S1 and Flood-S1
(mIoU, \%; mean $\pm$ std over 5 seeds). The BEiT-inherited default
used throughout the paper is $\lambda=0.75$; this column is the production
run (V100), whereas the other columns were run on an NVIDIA RTX A6000 GPU.}
\label{tab:llrd_lambda_sweep}

\footnotesize
\begin{tabular}{lcccc}
\toprule
$\lambda$ & 0.50 & 0.65 & 0.75 (default) & 0.90 \\
\midrule
DFC2020-S1 (mIoU)
& $56.11 \pm 0.13$
& $56.06 \pm 0.10$
& $56.00 \pm 0.08$
& $55.81 \pm 0.09$\textsuperscript{*} \\

Flood-S1 (mIoU)
& $80.09 \pm 0.24$
& $80.18 \pm 0.22$
& $80.21 \pm 0.10$
& $80.01 \pm 0.15$ \\
\bottomrule
\end{tabular}

\vspace{2pt}

\parbox{\columnwidth}{%
\scriptsize
\textsuperscript{*}Significantly lower than
$\lambda=0.75$ (paired $t$-test, $p_{\mathrm{BH}}=0.040$
across a family of 6 comparisons); no other pairwise contrast
reaches significance after correction.
}
\end{table}

We also test the sensitivity of the downstream task accuracy to 
the LLRD decay factor $\lambda$ by sweeping 
$\lambda \in \{0.5, 0.65, 0.75, 0.90\}$ on DFC2020-S1 and Flood-S1 
(Table~\ref{tab:llrd_lambda_sweep}). On Flood-S1, performance is 
statistically indistinguishable across the full range tested 
(repeated-measures ANOVA, $F(3,12)=1.44$, $p=0.28$), 
while on DFC2020-S1, we detect a small but consistent effect 
of $\lambda$ ($F(3,12)=22.6$, $p<0.001$). 
However, our default $\lambda=0.75$ is statistically indistinguishable 
from the best-performing value tested ($\lambda=0.5$; paired $t$-test, 
$p_{\mathrm{BH}}=0.19$), and the only significant pairwise contrast 
runs in the direction of $\lambda=0.75$ outperforming less aggressive 
decay ($\lambda=0.90$; $p_{\mathrm{BH}}=0.040$), not the reverse.
This pattern is consistent with the 
degradation at large $\lambda$ (near 1) for 
LLRD fine-tuning of vision transformers~\citep{dong2022clip},
and supports retaining the BEiT-inherited default of 
$\lambda=0.75$ without further tuning.

\subsubsection{Flood-S1 Ceiling Analysis}
\label{sec:ablation_flood}
The limited Flood-S1 gain ($+1.4$\,pp over no-pretrain baseline) persists
across all configurations in Table~\ref{tab:main}.
To investigate whether this ceiling is attributable to the loss function
or fusion strategy rather than pretraining, we conduct additional
ablations.

Table~\ref{tab:ablation_flood} reports Flood-S1 mIoU for different loss
function combinations and pre/post feature fusion strategies,
using the ConvNeXt-RS encoder pretrained with the ConvNeXt-512 decoder.
Every configuration is reported as $\text{mean}\pm\text{std}$ over 5 seeds
(42, 123, 456, 789, 1337), finetuned with the primary protocol
($lr{=}10^{-3}$, OneCycleLR with final\_div\_factor$\,{=}\,$100, except for the
CosineAnneal row) on an NVIDIA RTX A6000 GPU. Note that these results use the
ConvNeXt-512 pretrained encoder checkpoint and are therefore not directly
comparable to the UNet-128 pretrained results in Table~\ref{tab:main}.

Relative to the primary Subtraction+CE configuration ($80.11\pm0.15$), no
alternative gives a large improvement. Concatenation fusion with CE gives the
highest mean ($80.53\pm0.15$, $+0.42$\,pp; paired $t$-test, $p_{BH}{=}0.003$,
BH correction across the five comparisons), and Concatenation with focal loss
is similar ($80.52\pm0.27$, $+0.41$\,pp; $p_{BH}{=}0.054$). CosineAnneal
scheduling ($80.10\pm0.07$) and CE$+$TV ($80.20\pm0.12$, $p_{BH}{=}0.17$)
are indistinguishable from the baseline. Tversky loss alone is markedly worse
and unstable at this learning rate ($75.75\pm1.78$; one seed reached only
72.59; $p_{BH}{=}0.013$). Changes at the task head therefore move Flood-S1
mIoU by at most about 0.4\,pp, small next to the $+1.4$\,pp gain from
pretraining.

We suspect that the ceiling reflects the combination of limited training data
(3,000 samples), severe class imbalance (85.3\% non-flood pixels), and
a mismatch between single-image MIM pretraining objectives
and the temporal reasoning required for change detection; the ablation does
not isolate these factors.

\begin{table}[ht]
\caption{Flood-S1 ablation: finetuning configuration variants
(ConvNeXt-RS with ConvNeXt-512 pretrained encoder; mean$\pm$std over 5 seeds
(42, 123, 456, 789, 1337), $lr{=}10^{-3}$, NVIDIA RTX A6000 GPU).
CE: cross-entropy; TV: Tversky loss ($\alpha{=}0.3$, $\beta{=}0.7$);
Focal: focal loss ($\gamma{=}2$). Subtraction: $f_\text{post} - f_\text{pre}$;
Concatenation: $[f_\text{pre}; f_\text{post}]$.
Baseline (bold) is the primary configuration used throughout this paper
(its V100 production run in Table~\ref{tab:main} gives $80.11\pm0.10$).
All rows use OneCycleLR except the CosineAnneal row.
}
\label{tab:ablation_flood}
\centering
\scriptsize
\begin{tabular}{llllc}
\toprule
\textbf{LLRD} & \textbf{Scheduler} & \textbf{Fusion} & \textbf{Loss}
  & \textbf{Flood-S1 mIoU (\%)} \\
\midrule
No & OneCycleLR & Subtraction & CE & $79.99\pm0.14$ \\
\textbf{Yes} & \textbf{OneCycleLR} & \textbf{Subtraction} & \textbf{CE} & $\mathbf{80.11\pm0.15}$ \\
Yes & CosineAnneal & Subtraction & CE          & $80.10\pm0.07$ \\
Yes & OneCycleLR & Concatenation & CE + TV     & $80.20\pm0.12$ \\
Yes & OneCycleLR & Concatenation & Tversky     & $75.75\pm1.78$ \\
Yes & OneCycleLR & Concatenation & CE          & $80.53\pm0.15$ \\
Yes & OneCycleLR & Concatenation & Focal       & $80.52\pm0.27$ \\
\bottomrule
\end{tabular}
\end{table}

\subsection{Pretraining Data Scaling}
\label{sec:scaling}

\begin{figure*}[t]
\centering
\includegraphics[width=\linewidth]{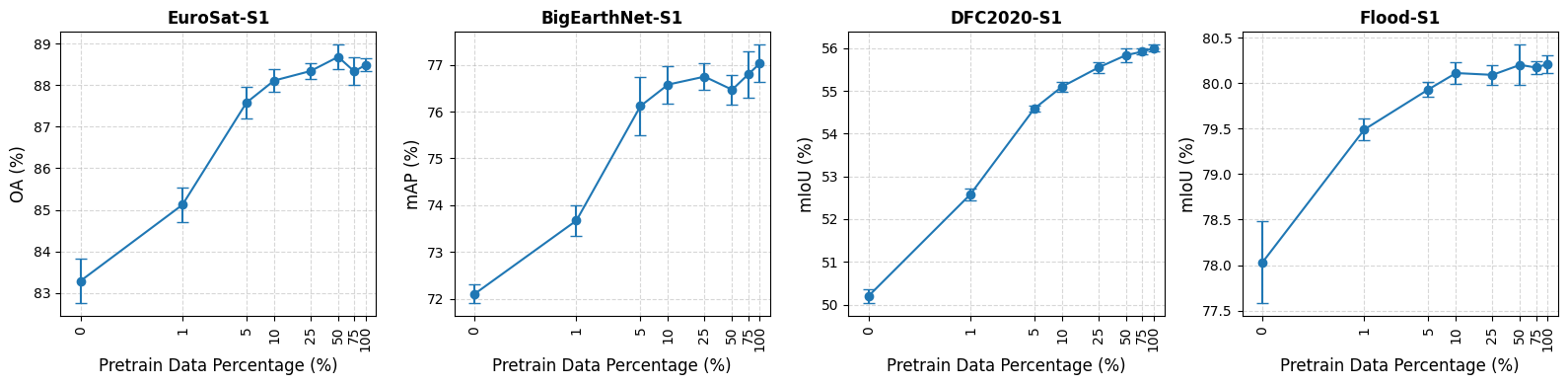}
\caption{Pretraining data scaling: finetuning accuracy vs.\ pretraining
data fraction (mean$\,\pm\,$std, 5 seeds), corresponding to
Table~\ref{tab:scaling}.}
\label{fig:scaling}
\end{figure*}

We evaluate the effect of pretraining data quantity by training
ConvNeXt-RS with UNet-128 on random subsets of SSL4EO-S12 v1.1
ranging from 1\% to 100\% (2,439 to 243,968 tiles), using a fixed
random subset seed (seed=0, framework default) for reproducibility.
All conditions are finetuned with LLRD and evaluated over 5 seeds.
Table~\ref{tab:scaling} and Figure~\ref{fig:scaling} report the 
results across all conditions, with
the 0\% condition (random initialization) serving as the lower bound.
For consistency with the scaling experiments, the 0\% condition is
also finetuned with LLRD. In contrast, the primary no-pretraining
baseline in Table~\ref{tab:main} is finetuned without LLRD.

\begin{table*}[ht]
\caption{Pretraining data scaling results (ConvNeXt-RS, UNet-128, LLRD,
mean$\,\pm\,$std, 5 seeds). $\ast$ indicates statistically significant
difference from 100\% (paired two-tailed t-test, BH FDR
correction across 28 comparisons, $p_\mathrm{BH} < 0.05$).}
\label{tab:scaling}
\centering
\small
\begin{tabular}{lrccccc}
\toprule
\textbf{Subset} & \textbf{Tiles} & \textbf{EuroSAT-S1}
  & \textbf{BigEarthNet-S1} & \textbf{DFC2020-S1} & \textbf{Flood-S1} \\
\midrule
0\% (random) & 0
  & $83.29{\pm}0.54^\ast$ & $72.10{\pm}0.20^\ast$
  & $50.20{\pm}0.17^\ast$ & $78.03{\pm}0.45^\ast$ \\
1\%  & 2,439
  & $85.12{\pm}0.42^\ast$ & $73.67{\pm}0.33^\ast$
  & $52.58{\pm}0.14^\ast$ & $79.49{\pm}0.12^\ast$ \\
5\%  & 12,198
  & $87.58{\pm}0.38^\ast$ & $76.12{\pm}0.62$
  & $54.59{\pm}0.07^\ast$ & $79.93{\pm}0.08^\ast$ \\
10\% & 24,396
  & $88.12{\pm}0.27$ & $76.58{\pm}0.40$
  & $55.10{\pm}0.12^\ast$ & $80.11{\pm}0.12$ \\
25\% & 60,992
  & $88.34{\pm}0.19$ & $76.75{\pm}0.29$
  & $55.55{\pm}0.13^\ast$ & $80.09{\pm}0.11$ \\
50\% & 121,984
  & $88.68{\pm}0.29$ & $76.47{\pm}0.32^\ast$
  & $55.84{\pm}0.16$ & $80.20{\pm}0.22$ \\
75\% & 182,976
  & $88.33{\pm}0.33$ & $76.80{\pm}0.50$
  & $55.92{\pm}0.08$ & $80.17{\pm}0.07$ \\
100\% & 243,968
  & $88.49{\pm}0.16$ & $77.03{\pm}0.40$
  & $56.00{\pm}0.08$ & $80.21{\pm}0.10$ \\
\bottomrule
\end{tabular}
\end{table*}

Paired two-tailed t-tests (N${=}5$ seeds, BH FDR
correction~\citep{benjamini1995controlling} across 28 comparisons) 
reveal task-dependent saturation behavior.
EuroSAT-S1 and Flood-S1 show no statistically significant difference
from the 100\% baseline for all conditions ${\geq}10\%$
($p_\mathrm{BH}{=}0.076$ and $p_\mathrm{BH}{=}0.402$ respectively).
BigEarthNet-S1 shows no significant difference at 25\%
($p_\mathrm{BH}{=}0.269$), though a non-monotonic dip at 50\%
(mean${=}76.47$ vs.\ $76.75$ at 25\%) yields a significant result
at that fraction ($p_\mathrm{BH}{=}0.005$), consistent with the
high inter-seed variance ($\sigma{\approx}0.50$\,pp) for this task;
discounting this single noisy condition, the effective saturation is
consistent with ${\geq}25\%$ of the pretraining corpus.
Semantic segmentation (DFC2020-S1) continues to improve significantly
up to 25\% ($p_\mathrm{BH}{=}0.002$), saturating between 25--50\%
($p_\mathrm{BH}{=}0.190$ at 50\%).
These results indicate that spatial representation quality in SAR
amplitude imagery is more sensitive to pretraining data quantity than
global semantic features.
We note that non-significance with $N{=}5$ is consistent with, but
does not prove, equivalence; results should be interpreted alongside the
effect sizes in Table~\ref{tab:scaling}.

\subsection{Finetuning Label Efficiency}
\label{sec:label_efficiency}

\begin{figure*}[t]
\centering
\includegraphics[width=\linewidth]{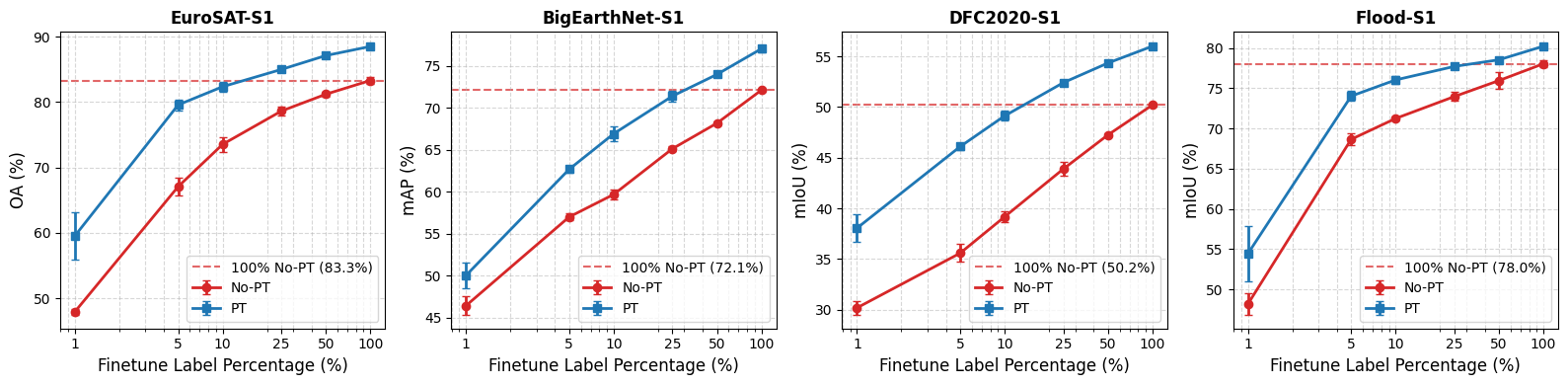}
\caption{Label efficiency: finetuning accuracy vs.\ labeled-data
fraction, pretrained (PT) vs.\ random initialization (No-PT)
(mean$\,\pm\,$std, 5 seeds), corresponding to Table~\ref{tab:labeleff}.
Dashed line marks the 100\%-label No-PT level.}
\label{fig:labeleff}
\end{figure*}

To assess whether SparK-SAR pretraining reduces reliance on labeled 
data, we finetune the primary configuration (UNet-128 decoder, LLRD) 
on 1\%, 5\%, 10\%, 25\%, and 50\% subsets of each task's training 
labels, comparing pretrained initialization against random initialization 
at each fraction (5 seeds each, with the same seed controlling both 
model initialization and label subset). Table~\ref{tab:labeleff} 
and Figure~\ref{fig:labeleff}
report the results alongside the existing 100\%-label result from 
Table~\ref{tab:main}.

Pretraining provides a statistically significant advantage over 
random initialization at every label fraction and every task 
(paired $t$-test, BH-corrected across the 20 comparisons; 
all $p_{BH} < 0.05$, ranging from $7.2\times10^{-7}$ to 
$1.07\times10^{-2}$), including the most extreme setting we test: 
30 labeled samples for Flood-S1. For EuroSAT-S1, BigEarthNet-S1, 
and DFC2020-S1, the absolute gain from pretraining peaks at a modest 
5--10\% label budget and recedes as labeled data becomes abundant; 
Flood-S1 instead shows its largest advantage at the very smallest 
label fraction (1\%), decaying monotonically thereafter. Practically, 
pretrained models reach the random-initialization 100\%-label result 
using only 10--25\% of the labels for EuroSAT-S1 and DFC2020-S1, 
and 25--50\% for BigEarthNet-S1 and Flood-S1, cutting the labeling 
burden required to match full-data performance by roughly half to 
three-quarters.

\begin{table*}[t]
\centering
\caption{Label efficiency: finetuning accuracy vs.\ labeled-data fraction, 
pretrained (PT) vs.\ random initialization (No-PT), UNet-128 decoder, 
finetuned using LLRD, 5 seeds. All pretrained $>$ random initialized gaps 
are statistically significant (paired $t$-test, BH-corrected across 20 
comparisons, $p_{BH}<0.05$; max $p_{BH}=0.011$).}
\label{tab:labeleff}
\scriptsize
\begin{tabular}{lcccccccc}
\toprule
& \multicolumn{2}{c}{EuroSAT-S1 (OA\%)} & \multicolumn{2}{c}{BigEarthNet-S1 (mAP\%)} & \multicolumn{2}{c}{DFC2020-S1 (mIoU\%)} & \multicolumn{2}{c}{Flood-S1 (mIoU\%)} \\
\cmidrule(lr){2-3}\cmidrule(lr){4-5}\cmidrule(lr){6-7}\cmidrule(lr){8-9}
Labels & No-PT & PT & No-PT & PT & No-PT & PT & No-PT & PT \\
\midrule
1\% & 47.94$\pm$0.49 & 59.54$\pm$3.65 & 46.41$\pm$1.10 & 50.01$\pm$1.52 & 30.19$\pm$0.70 & 38.05$\pm$1.39 & 48.17$\pm$1.36 & 54.43$\pm$3.38 \\
5\% & 67.07$\pm$1.31 & 79.56$\pm$0.83 & 56.99$\pm$0.40 & 62.67$\pm$0.36 & 35.59$\pm$0.87 & 46.10$\pm$0.33 & 68.64$\pm$0.69 & 74.01$\pm$0.60 \\
10\% & 73.53$\pm$1.13 & 82.34$\pm$0.80 & 59.67$\pm$0.55 & 66.90$\pm$0.87 & 39.18$\pm$0.55 & 49.13$\pm$0.50 & 71.23$\pm$0.28 & 76.02$\pm$0.46 \\
25\% & 78.62$\pm$0.70 & 84.98$\pm$0.35 & 65.12$\pm$0.30 & 71.39$\pm$0.62 & 43.90$\pm$0.70 & 52.41$\pm$0.29 & 73.96$\pm$0.55 & 77.70$\pm$0.36 \\
50\% & 81.19$\pm$0.44 & 87.10$\pm$0.30 & 68.16$\pm$0.20 & 73.97$\pm$0.18 & 47.25$\pm$0.13 & 54.34$\pm$0.17 & 75.94$\pm$1.04 & 78.52$\pm$0.32 \\
100\%$^*$ & 83.29$\pm$0.54 & 88.49$\pm$0.16 & 72.10$\pm$0.20 & 77.03$\pm$0.40 & 50.20$\pm$0.17 & 56.00$\pm$0.08 & 78.03$\pm$0.45 & 80.21$\pm$0.10 \\
\bottomrule
\end{tabular}
\\[2pt]
{\footnotesize $^*$The PT column at 100\% matches Table~\ref{tab:main} 
(SparK-SAR + UNet-128, LLRD). The No-PT column applies LLRD for consistency 
with lower label fractions; the no-pretraining baseline in 
Table~\ref{tab:main} does not use LLRD.}
\end{table*}

\subsection{Representation Analysis}
\label{sec:cka}
To complement the downstream performance analysis, we compute
Centered Kernel Alignment (CKA)~\citep{kornblith2019similarity} between
encoder representations at different pretraining data fractions.
CKA measures linear representational similarity between two neural
networks on a shared set of inputs.

We extract Stage~3 post global average pooling features ($[N,384]$)
from all eight pretrained encoders (0--100\%) on each Copernicus-Bench
S1GRD test set, and compute pairwise unbiased linear CKA with Z-score
feature standardization~\citep{nguyen2021wide} to mitigate channel
variance skew.

We use Stage~3 rather than the final Stage~4 representation because GAP
over the $14{\times}14$ Stage~4 output would collapse the spatial
structure that our resolution ablation specifically shows to be valuable,
whereas Stage~4 features retained at full resolution ($[N,14,14,768]$)
have no natural scalar CKA summary. Stage~3 avoids this collapse while
still providing a compact, fixed-size representation for similarity
comparison. This choice is also consistent with recent evidence that
task-relevant information is often more accessible in intermediate
layers than in final-layer embeddings across several self-supervised
remote sensing architectures~\citep{romero2026transfer}.

\begin{figure*}[ht]
\centering
\includegraphics[width=\linewidth]{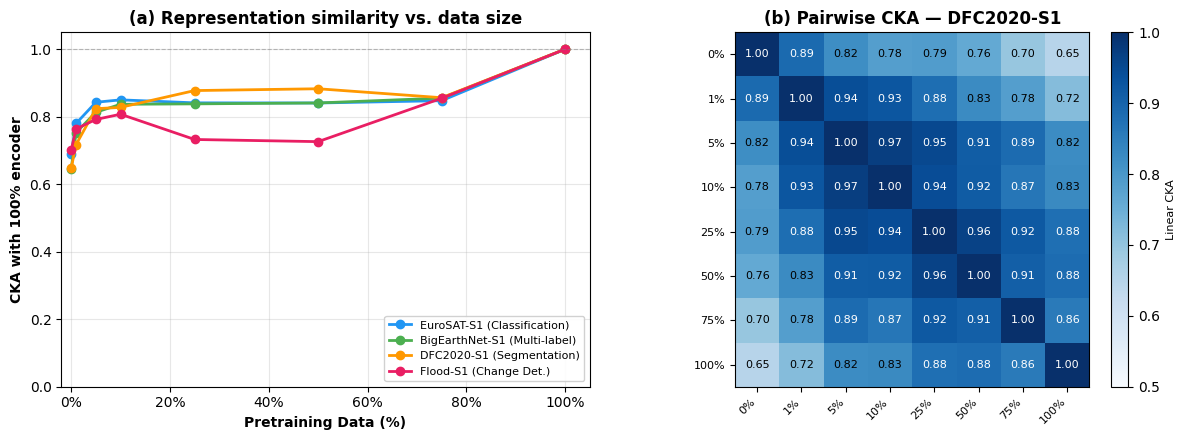}
\caption{(a) CKA similarity to fully pretrained (100\%) encoder vs.\
pretraining data fraction (Stage~3 post-GAP features, unbiased HSIC,
Z-score standardization). (b) Pairwise CKA heatmap for DFC2020-S1.
Values near the diagonal confirm that representations converge as data
increases, with saturation consistent with downstream performance.}
\label{fig:cka}
\end{figure*}

Figure~\ref{fig:cka} shows CKA similarity to the fully pretrained
(100\%) encoder as a function of pretraining data fraction.
For EuroSAT-S1, BigEarthNet-S1, and DFC2020-S1, CKA increases
near monotonically from random initialization (0.64--0.70) to plateau
values of 0.83--0.88, broadly consistent with the downstream
performance saturation observed in Section~\ref{sec:scaling}.

We observe an anomalous dip in CKA similarity for Flood-S1 at the 
25\% and 50\% pretraining-data fractions (0.733 and 0.726, respectively, 
versus 0.807 at 10\% and 0.855 at 75\%). We do not have a confident 
explanation for this non-monotonicity and report it as an open observation
rather than speculate on its cause. It does not affect the downstream 
accuracy trend (Table~\ref{tab:scaling}), which remains monotonic within 
noise across the same fractions.

A non-trivial CKA baseline of ${\approx}0.65$--$0.70$ at 0\%
(random initialization) reflects the strong inductive bias imposed by
ConvNeXt's depthwise convolutional structure, which generates
spatially correlated SAR features even without training.

\subsection{Geographic Data Overlap Analysis}
\label{sec:leakage}
SSL4EO-S12 v1.1 and the Copernicus-Bench evaluation sets both draw
from the Sentinel-1 archive and share similar global geographic
sampling strategies, raising the possibility of geographic overlap
between pretraining and evaluation data.
To assess whether our reported gains are inflated by such overlap,
we identify pretraining tiles with geographic proximity to test set
samples using task-specific thresholds derived from the physical patch
dimensions of each dataset pair, given a Sentinel-1 GSD of
10\,m/pixel: 1,640\,m for EuroSAT-S1, 1,920\,m for BigEarthNet-S1,
and 2,440\,m for Flood-S1. DFC2020-S1 is not assessed due to the
absence of coordinate metadata in the dataset release.
Under these thresholds, 10.9\% of EuroSAT-S1 test samples, 4.3\% of
BigEarthNet-S1 test samples, and 10.1\% of Flood-S1 test samples have
at least one geographically proximate pretraining tile; from the
pretraining side, only 1,406 of 243,968 SSL4EO-S12 v1.1 tiles (0.58\%)
are implicated across all three assessable tasks combined.

To directly test whether this overlap inflates our reported gains, we
re-pretrain SparK-SAR from scratch on a de-collided version of
SSL4EO-S12 v1.1 with all 1,406 overlapping tiles removed (242,562
remaining), and re-finetune on all four downstream tasks (5 seeds
each). The de-collided model was pretrained (FP16 mixed precision, global batch
size 256) on a single NVIDIA A100 80\,GB GPU
and finetuned (FP32) on single NVIDIA RTX A6000 GPUs, whereas the original models were
pretrained on 4$\times$V100 32\,GB GPUs (64 samples per GPU) and finetuned on V100 GPUs;
the remainder of the recipe is unchanged, so hardware and per-device batch size are
minor additional differences between the two columns of Table~\ref{tab:decollision}.
Table~\ref{tab:decollision} reports the results: all four tasks
change by less than 0.1 percentage points relative to the original
pretraining, within one standard deviation, confirming that the
reported gains are not attributable to geographic leakage. DFC2020-S1
remains an open limitation for the overlap analysis itself, since its
dataset release provides no coordinate metadata to check test-set
proximity directly; its unchanged performance under the de-collided
checkpoint above offers indirect reassurance only.

\begin{table}[ht]
\caption{Full experimental de-collision: SparK-SAR re-pretrained on
SSL4EO-S12 v1.1 with all overlapping tiles removed, re-finetuned on
all four tasks (mean$\,\pm\,$std, 5 seeds).}
\label{tab:decollision}
\centering
\begin{tabular}{lccc}
\toprule
\textbf{Task} & \textbf{Original} & \textbf{De-collided} & $\Delta$ \\
\midrule
EuroSAT-S1     & 88.49$\pm$0.16 & 88.46$\pm$0.31 & $-0.03$ \\
BigEarthNet-S1 & 77.03$\pm$0.40 & 77.08$\pm$0.47 & $+0.05$ \\
DFC2020-S1     & 56.00$\pm$0.08 & 56.02$\pm$0.10 & $+0.02$ \\
Flood-S1       & 80.21$\pm$0.10 & 80.30$\pm$0.18 & $+0.09$ \\
\bottomrule
\end{tabular}
\end{table}

\subsection{Qualitative Results}
\label{sec:qualitative}
\begin{figure*}[ht]
\centering
\includegraphics[width=\linewidth]{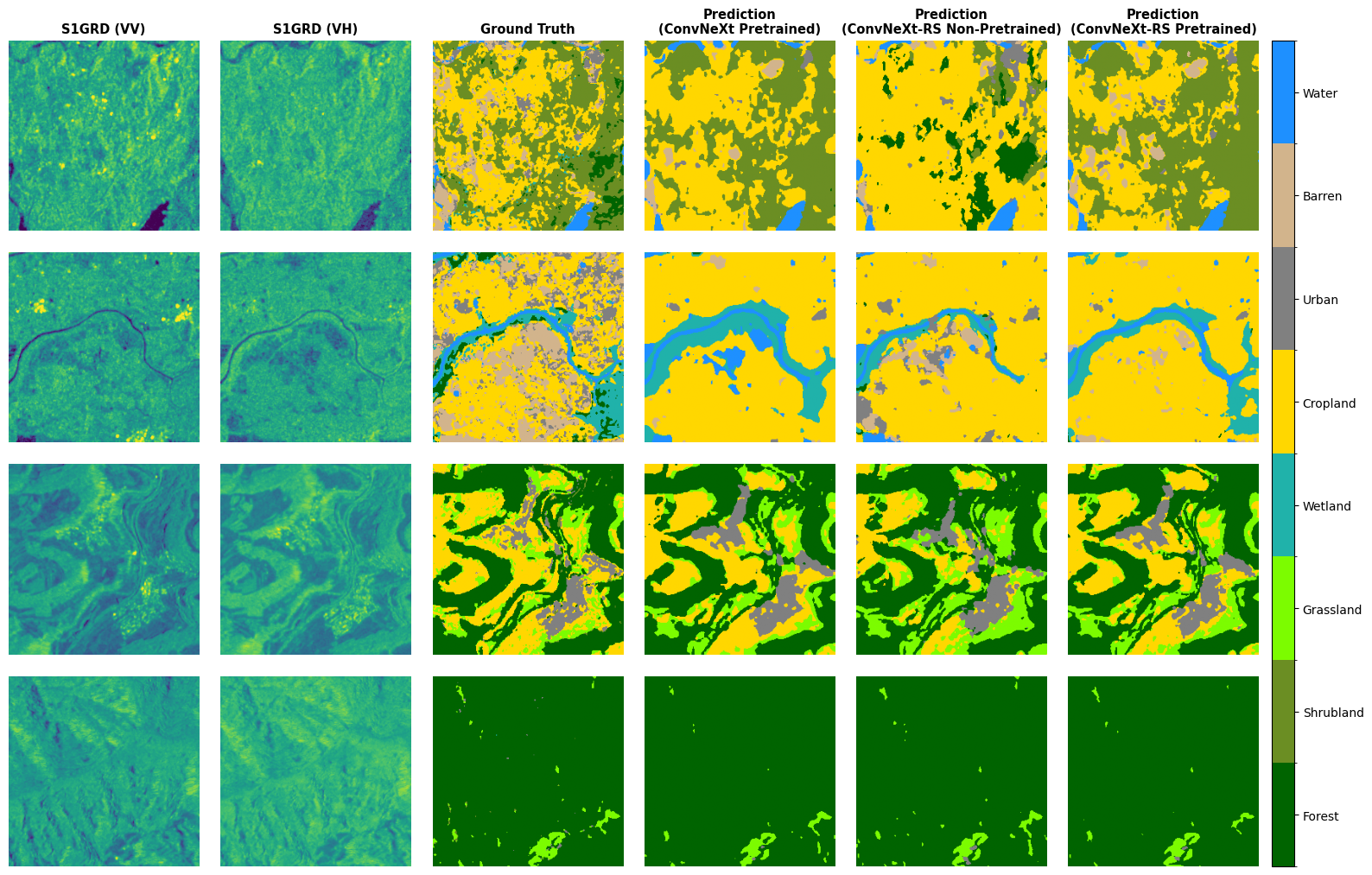}
\caption{DFC2020-S1 qualitative comparison across four representative
scenes (rows). Columns from left to right: VV input, VH input, ground
truth, and predictions from the standard-resolution ConvNeXt encoder
(pretrained, no LLRD; Table~\ref{tab:ablation_lr}), the ConvNeXt-RS
encoder without pretraining (no LLRD; Table~\ref{tab:main}), and our
finetuned ConvNeXt-RS encoder (pretrained, LLRD; Table~\ref{tab:main}).
LLRD was evaluated only for ConvNeXt-RS.}
\label{fig:dfc2020}
\end{figure*}

Figure~\ref{fig:dfc2020} compares four representative DFC2020-S1 scenes
across three configurations reported in this paper:
the standard-resolution ConvNeXt encoder (pretrained, no LLRD;
Table~\ref{tab:ablation_lr}),
the ConvNeXt-RS encoder without pretraining (no LLRD; Table~\ref{tab:main}),
and our finetuned ConvNeXt-RS encoder (pretrained, LLRD; Table~\ref{tab:main}).

The first row presents a mixed landscape dominated by shrubland and
cropland, with smaller barren, urban, and forest patches and a small
water body.
All three predictions broadly match the ground truth, but notable
differences are already apparent: the standard ConvNeXt encoder
over-predicts cropland in the central region and recovers almost none
of the forest; the non-pretrained ConvNeXt-RS encoder collapses most of
the shrubland into cropland, introduces spurious forest in areas the
ground truth labels shrubland, and suppresses the barren regions; in
contrast, the pretrained ConvNeXt-RS encoder recovers the largest share
of the ground truth's shrubland and retains the most barren land.

In the second row, a cropland-dominated scene containing a prominent water body, 
all three configurations accurately capture the water feature and surrounding cropland. 
However, both the standard-resolution ConvNeXt encoder and the non-pretrained ConvNeXt-RS 
encoder make spurious predictions of other classes in the region where the ground truth 
shows barren land, whereas the pretrained ConvNeXt-RS encoder avoids these spurious 
cross-class predictions but instead over-extends cropland into the barren region.

The third row is the most heterogeneous scene, mixing forest, cropland,
grassland, and urban patches.
All three configurations recover the overall spatial layout, though
predictions are smoother than the ground truth and suppress fine-grained
class boundaries; urban patches are captured at broadly similar
locations across all three models.
The main variation again separates the non-pretrained encoder from the
two pretrained configurations: it reassigns a substantial part of the
ground truth's cropland to grassland, whereas both pretrained
configurations preserve the cropland-grassland boundary more closely.

In the fourth row, a largely forested scene with scattered grassland,
all three predictions are predominantly dark green and closely resemble
the ground truth, differing mainly in that the non-pretrained encoder
places more of its grassland predictions outside the ground truth's
grassland patches than either pretrained configuration.

\begin{figure*}[ht]
\centering
\includegraphics[width=\linewidth]{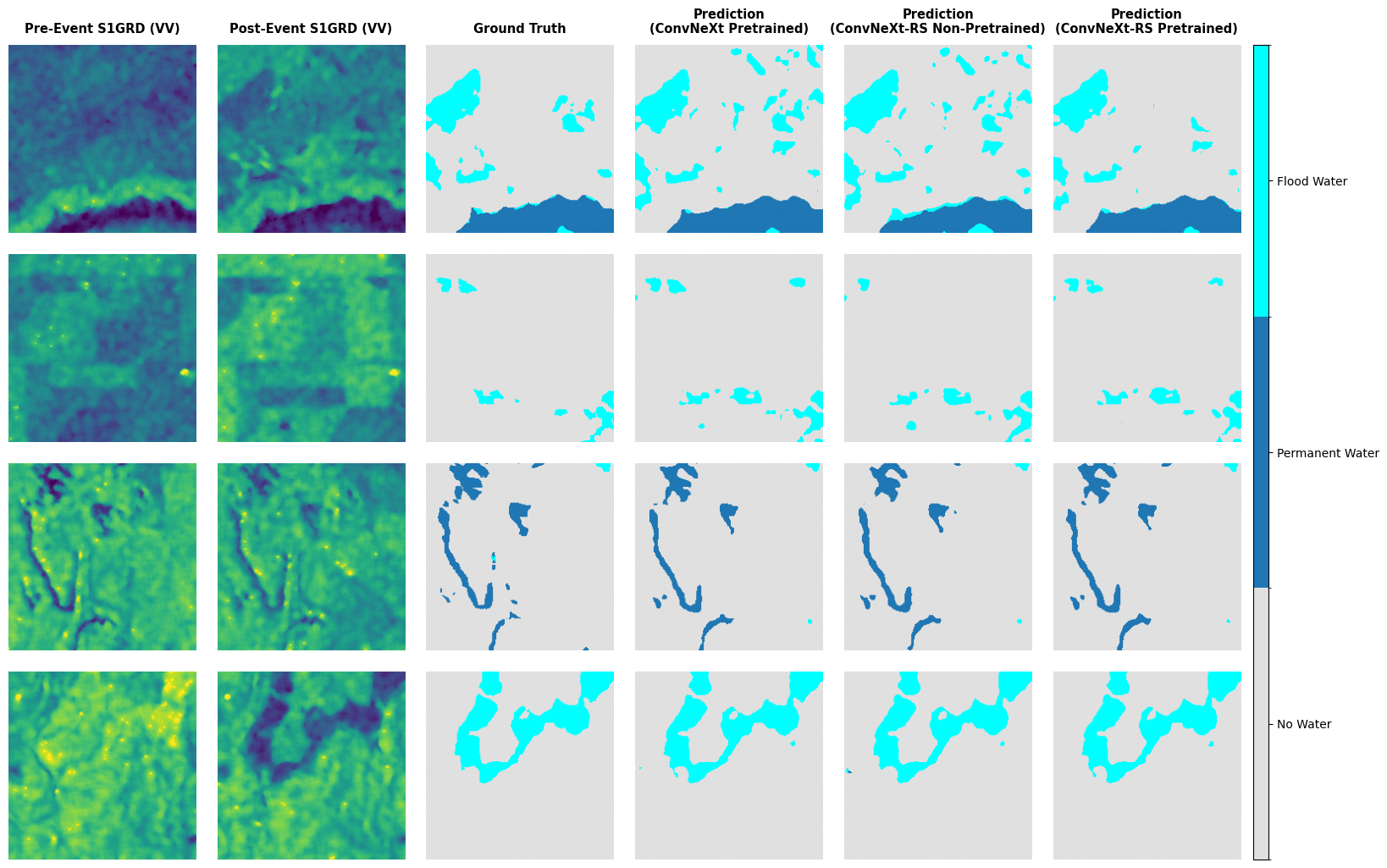}
\caption{Flood-S1 qualitative comparison across four representative
scenes (rows). Columns from left to right: pre-flood VV, post-flood VV,
ground truth, and predictions from the standard-resolution ConvNeXt
encoder (pretrained, no LLRD; Table~\ref{tab:ablation_lr}), the
ConvNeXt-RS encoder without pretraining (no LLRD; Table~\ref{tab:main}),
and our finetuned ConvNeXt-RS encoder (pretrained, LLRD;
Table~\ref{tab:main}). LLRD was evaluated only for ConvNeXt-RS.}
\label{fig:flood}
\end{figure*}

Figure~\ref{fig:flood} shows the same three-way comparison for four
representative Flood-S1 scenes.
In the first row, where the ground truth contains a large contiguous
flood-water region in the upper portion and a permanent-water body in
the lower portion, both pretrained configurations label the
permanent-water body almost entirely correctly, whereas the
non-pretrained ConvNeXt-RS encoder assigns a noticeable fraction of it
to flood water.
Both the standard-resolution ConvNeXt encoder and the non-pretrained
ConvNeXt-RS encoder additionally make multiple spurious predictions of
flood water in the top-right portion of the scene, whereas the
pretrained ConvNeXt-RS encoder makes none there.

The second row presents a more challenging, spatially sparse case with
small scattered flood-water patches against an otherwise water-free
background; all three configurations detect the approximate locations
of most patches, but the non-pretrained encoder misses noticeably more
of the flooded area than either pretrained configuration.

The third row contains mainly permanent water in the form of a
prominent meandering channel, whose overall course all three
configurations recover at comparable accuracy.

In the fourth row, flood-water patches are distributed across the
scene; all three configurations recover the main flooded areas, with
variation confined to the finer details of individual patch boundaries
and smaller isolated regions.
Taken together, the clearest and most consistent difference across
these scenes is the non-pretrained encoder's weaker separation of flood
water from both permanent water and dry land.

\section{Discussion}
\label{sec:discussion}
Of the three design changes evaluated, the higher-resolution
ConvNeXt-RS encoder matters most for dense prediction: its gains are
largest on segmentation and change detection, which is exactly the
signal we would expect if the standard stride-$32{\times}$ ConvNeXt is
simply too coarse for SAR pixel-level tasks. Halving the stem stride
fixes this without adding a single parameter.

The decoder ablation (Section~\ref{sec:ablation_decoder}) tells a
complementary story: UNet-128 and ConvNeXt-512 land at statistically
indistinguishable downstream accuracy ($p{=}0.17$), so whatever the
original SparK recipe's hierarchical decoder buys on natural images,
it is not required to transfer well to S1GRD at a matched learning
rate. That makes ConvNeXt-512 the better default whenever VRAM is
tight, since it recovers 7.5--8.5\% per GPU (hardware-dependent) for
no accuracy cost. LLRD, meanwhile, is a smaller but broadly positive
lever for the UNet-128-pretrained encoder: mean performance improves and
variance shrinks across all four tasks, reaching significance on two of them
after correction (Section~\ref{sec:ablation_llrd}), enough that we recommend
it as a default for pretrained encoders rather than an optional extra; it did
not help a non-pretrained encoder (Section~\ref{sec:finetune}).

The data scaling analysis reveals that SAR MIM pretraining is
remarkably data-efficient: classification tasks and change detection
saturate with as few as 10--25\% of the pretraining corpus (24,396
tiles), while dense semantic segmentation (DFC2020-S1) keeps improving
significantly up to 25\% (60,992 tiles) and continues to gain, more
gradually, through 50\% (121,984 tiles).
This saturation behavior suggests that the limited visual diversity 
of SAR amplitude imagery, which is typically dominated by a relatively 
small number of backscatter regimes corresponding to major land-cover 
types, can be effectively characterized by a surprisingly small 
pretraining corpus.
Consequently, researchers need not curate massive SAR datasets to
benefit from the SparK-SAR pretraining method; a geographically diverse
collection of $\sim$25,000 tiles may be sufficient for most
applications.

The CKA representation analysis (Section~\ref{sec:cka}) corroborates
the downstream saturation findings: encoder representations converge
toward the fully pretrained configuration with increasing data, and
this convergence broadly mirrors the performance plateau observed in
Table~\ref{tab:scaling}.
The non-trivial CKA baseline at random initialization reflects the
strong spatial inductive bias of ConvNeXt on SAR data, an observation
with implications for future architectural choices in SAR SSL.

During development of SparK-SAR, several alternative configurations
were explored but found to provide no consistent benefit.

VICReg regularization~\citep{bardes2022vicreg} increased representation
dimensionality but degraded downstream performance, consistent with a
conflict between the generative SparK reconstruction objective and the
decorrelation objective of VICReg.

Fisher-Tippett-based SAR-specific reconstruction 
loss~\citep{fisher1928, guan2019} and distillation from
a cross-modal optical teacher (DINOv3-style~\citep{simeoni2026dinov3}) 
produced reduced training
losses but no downstream gains, suggesting that representations aligned
with optical imagery are not directly transferable to SAR tasks.

These negative results highlight the importance of domain-specific
objective design for SAR SSL and are documented here to prevent
redundant exploration by the community.

A dual-branch polarimetric stem processing VV and VH channels
separately before $1{\times}1$ projection fusion was also evaluated,
but did not result in improvement on classification tasks and
regressed on dense prediction ($-1.44$\,pp on DFC2020-S1, $-0.52$\,pp on
Flood-S1), suggesting that joint polarimetric processing in a shared
convolutional stem is preferable for S1GRD MIM pretraining.

Flood-S1's comparatively modest $+1.4$\,pp gain has a natural
explanation once we consider what the pretraining objective actually
teaches the encoder. Change detection asks the model to reason
jointly over a pre-event and a post-event acquisition, but MIM
pretraining, single-image by construction, never exposes the encoder
to any notion of temporal difference at all. An encoder that has only
ever learned to reconstruct one image at a time has correspondingly
little temporal structure to transfer. The Flood-S1 ablation
(Table~\ref{tab:ablation_flood}) shows that the loss functions and fusion
strategies we tried at the task head change Flood-S1 mIoU by at most about
0.4\,pp (Concatenation fusion with CE or focal loss), so the limitation looks
like it may sit upstream, in the
pretraining objective itself, which points toward multi-temporal
pretraining as the more promising direction for future work rather
than further task-head engineering.

BigEarthNet-S1 is the one task where SparK-SAR (\ms{77.03}{0.40})
trails, coming in marginally behind both Copernicus-FM
($77.9{\pm}0.0$\%) and SoftCon ($78.7{\pm}0.0$\%) under frozen
evaluation. We note that the Copernicus-Bench dataset wrappers pass
geographic coordinates and acquisition timestamps as metadata to
Copernicus-FM for BigEarthNet-S1, while DFC2020-S1 passes NaN
metadata in lieu of absent coordinate information~\citep{wang2025copernicus}.
In contrast, ConvNeXt-RS receives no metadata whatsoever, relying
solely on the two SAR amplitude channels (VV, VH) for all tasks.
One plausible contributor is that BigEarthNet-S1's labels come from the
CORINE Land Cover map~\citep{sumbul2019bigearthnet,eea2020corine}, and
European land-cover classes are not spread uniformly: they cluster
geographically. Combined with the real coordinates and acquisition
times Copernicus-FM receives as input, that clustering gives it a
location-based shortcut to lean on that our model, seeing only VV/VH
amplitude, simply does not have access to. This is consistent with
the task-level pattern: Copernicus-FM trails our model on DFC2020-S1
(where metadata is all-NaN) and Flood-S1 (where the label is
transient and event-specific, limiting the utility of location
priors), while its strongest advantage over our model occurs on
BigEarthNet-S1 (where location and time metadata are available and
labels are geographically clustered). SoftCon's strong BigEarthNet-S1
result ($78.7{\pm}0.0$\%) likely reflects its multi-label land-cover
guidance during pretraining~\citep{wang2024softcon}. We offer this as
a plausible explanation, not a demonstrated one.

These findings also carry practical implications for operational Earth
observation pipelines that must produce map products under real-world time
and annotation constraints. The label efficiency analysis 
(Section~\ref{sec:label_efficiency}) shows that SparK-SAR's advantage over 
a non-pretrained baseline on Flood-S1 is largest precisely at the most 
extreme low-label setting we test (1\% labeled samples, a +6.26 pp gain in 
mIoU over random initialization), the operating regime closest to a real 
rapid-response flood mapping scenario such as Copernicus Emergency Management 
Service (EMS) products, where annotated post-event imagery for a specific 
flood event is scarce or unavailable at the time a map is needed. 

Similarly, the pretraining data scaling results (Section~\ref{sec:scaling}) show that 
classification and change detection performance saturate with only 10--25\% 
of the full 243,968 tile pretraining corpus, indicating that agencies 
producing regional land-cover products in the style of CORINE or the ESA 
Climate Change Initiative (CCI) Land Cover map from Sentinel-1 archives 
need not assemble a global-scale SAR corpus to benefit from MIM pretraining. 

SparK-SAR pretraining is a one-time cost across four V100 32~GB GPUs, 
completing in under nine hours (Table~\ref{tab:compute}), while finetuning 
the resulting checkpoint across all four Copernicus-Bench tasks evaluated 
in this paper then completes in a further $\approx$18.6 GPU-hours per seed on a 
single V100 GPU, well within reach of operational and research groups 
without dedicated large-scale HPC infrastructure, a practical constraint 
that we view as at least as relevant to real-world adoption as the 
accuracy gains reported above.

The broader case our results build is a practical one: for SAR
remote sensing, a compute-efficient, domain-specific CNN
can be a real alternative to chasing ever-larger multimodal ViT
foundation models. A 27.8M-parameter encoder pretrained on one
modality, on four V100 32\,GB GPUs, in under a day, holds its own
against, and on most tasks beats, frozen encoders with five times the
parameters and orders of magnitude more pretraining data and compute
behind them.

\section{Conclusions}
\label{sec:conclusion}
This work investigates how masked self-supervised pretraining can be 
adapted to the particular characteristics of Sentinel-1 GRD imagery. 
The proposed SparK-SAR framework combines a high-resolution ConvNeXt-RS 
encoder producing $14{\times}14$ feature maps with lightweight 
reconstruction decoders designed to avoid the numerical difficulties 
observed during mixed-precision pretraining on SAR data. 
Together with layer-wise learning rate decay during finetuning, 
these modifications enable a 27.8M-parameter encoder to transfer 
effectively to four distinct Copernicus-Bench S1GRD downstream tasks 
within a limited computational budget.

The experiments also provide evidence that increasing pretraining data 
does not yield uniform benefits across downstream tasks. 
The 27.8M-parameter encoder exceeds the performance of frozen 
ViT-based encoders with as much as $5{\times}$ more parameters on 
three of the four evaluated tasks. 
Moreover, the data-scaling experiments show that classification
performance largely reaches saturation after using 10--25\% of the
SSL4EO-S12 v1.1 corpus, whereas dense prediction tasks continue to
improve significantly up to 25\% and keep gaining, more gradually,
through 50\% of the training set.
CKA analysis provides a complementary view of this behavior: 
as more data is introduced, the learned representations progressively
approach those obtained with the full pretraining set, following trends 
that are consistent with the observed downstream results.

Taken together, the findings indicate that effective SAR representation 
learning can be achieved without relying exclusively on very large-scale 
pretraining. Compact CNN models, when paired with objectives and 
architectures adapted to SAR characteristics, can provide useful 
representations with substantially lower computational requirements. 
At the same time, the task-dependent scaling behavior highlights limitations 
of single-image pretraining, particularly for temporal change analysis.
Future work will therefore investigate pretraining objectives that explicitly 
exploit multi-temporal observations, as well as extensions to interferometric 
SAR that incorporate phase and coherence alongside amplitude information.

\section*{Data Availability Statement}
The SSL4EO-S12 v1.1 dataset is publicly available
at \url{https://huggingface.co/datasets/embed2scale/SSL4EO-S12-v1.1}
\citep{blumenstiel2025ssl4eov11}. The Copernicus-Bench evaluation
benchmark is publicly available at
\url{https://huggingface.co/datasets/wangyi111/Copernicus-Bench}
\citep{wang2025copernicus}. Code and pretrained weights will be made
available upon acceptance.

\section*{Acknowledgment}
The authors thank Prof.\ Cham Tat Jen and Assoc.\ Prof.\ Benoit Taisne
(Thesis Advisory Committee) for their feedback during the course of
this work. 
Y.\ Natsume is supported by the Nanyang Technological University
Provost Graduate Award (NTU PGA).
Computational resources were provided by the College of Computing and
Data Science, Nanyang Technological University.

\section*{
Declaration of generative AI and AI-assisted technologies in the 
manuscript preparation process
}
During the preparation of this work, the authors used Claude (Anthropic) to assist with
drafting and language editing of the Introduction, Related Work, Materials and Methods,
Discussion, and Conclusion sections, including verification of citations and technical
statements against source material. The authors reviewed and edited the output as needed
and take full responsibility for the content of the published article. All experimental
design, code, data analysis, scientific content, and conclusions are the authors' own.

\bibliographystyle{elsarticle-harv}
\bibliography{references}

\end{document}